\documentclass[lettersize,journal]{IEEEtran}
\usepackage{amsmath,amsfonts}

\usepackage{algorithmic}
\usepackage{algorithm}
\usepackage{array}
\usepackage[caption=false,font=normalsize,labelfont=sf,textfont=sf]{subfig}
\usepackage{textcomp}
\usepackage{stfloats}
\usepackage{url}
\usepackage{verbatim}
\usepackage{graphicx}
\usepackage{cite}
\usepackage{xcolor}
\usepackage{soul}
\begin{document}

\newcommand{\edits}[1]{\textcolor{blue}{#1}}

\title{Soft Wrist Exosuit Actuated by \\Fabric Pneumatic Artificial Muscles}

\author{Katalin Schäffer, \textit{Graduate Student Member}, \textit{IEEE}, Yasemin Ozkan-Aydin, \textit{Member}, \textit{IEEE}, \\and Margaret M. Coad, \textit{Member}, \textit{IEEE}
\thanks{This work was supported by the Ministry of Culture and Innovation of Hungary from the National Research, Development and Innovation Fund, financed under the TKP2021-NKTA funding scheme (project no. TKP2021-NKTA-66).}
\thanks{Katalin Schäffer is with the Department of Aerospace and Mechanical Engineering, University of Notre Dame, Notre Dame IN 46556, USA, and also with the 
Faculty of Information Technology and Bionics, Pázmány Péter Catholic University, 1083 Budapest, Hungary (e-mail: kschaff2@nd.edu).}
\thanks{Yasemin Ozkan-Aydin is with the Department of Electrical Engineering, University of Notre Dame, Notre Dame IN 46556, USA (e-mail: yozkanay@nd.edu).}
\thanks{Margaret M. Coad is with the Department of Aerospace and Mechanical Engineering, University of Notre Dame, Notre Dame IN 46556, USA (e-mail: mcoad@nd.edu).}}

\markboth{Journal of \LaTeX\ Class Files,~Vol.~14, No.~8, August~2021}
{Shell \MakeLowercase{\textit{et al.}}: A Sample Article Using IEEEtran.cls for IEEE Journals}

\maketitle

\begin{abstract}
Recently, soft actuator-based exosuits have gained interest, due to their high strength-to-weight ratio, inherent safety, and low cost.
We present a novel wrist exosuit actuated by fabric pneumatic artificial muscles that has lightweight wearable components (160 g) and can move the wrist in flexion/extension and ulnar/radial deviation. We derive a model representing the torque exerted by the exosuit and demonstrate the use of the model to choose an optimal design for an example user.
We evaluate the accuracy of the model by measuring the exosuit torques throughout the full range of wrist flexion/extension. We show the importance of accounting for the displacement of the mounting points, as this helps to achieve the smallest mean absolute error (0.283 Nm) compared to other models.
Furthermore, we present the measurement of the exosuit-actuated range of motion on a passive human wrist.
Finally, we demonstrate the device controlling the passive human wrist to move to a desired orientation along a one and a two-degree-of-freedom trajectory.
The evaluation results show that, compared to other pneumatically actuated wrist exosuits, the presented exosuit is lightweight and strong (with peak torque of 3.3~Nm) but has a limited range of motion.
\end{abstract}


\section{Introduction}
\IEEEPARstart{U}{pper} limb wearable assistive devices can be useful in a variety of scenarios. Healthy people can benefit from physical assistance to avoid fatigue and muscle strain during repetitive tasks or to augment their natural physical abilities~\cite{shen2020upper}, and the physically impaired can benefit from assistance for activities of daily living~\cite{nam2019external}, or assistance and resistance during rehabilitation exercises~\cite{qassim2020review}. Especially over the last decade, soft wearable assistive devices, also known as soft exosuits, have attracted much interest from the research community. The term ``soft exosuits" is used as a collective term for wearable robotic devices made from compliant materials~\cite{thalman2020review}. They wrap around the user’s body and rely on its structural integrity instead of an additional rigid frame to transfer reaction forces between body segments~\cite{xiloyannis2021soft}.
Compared to their rigid counterparts where the joints are motorized, soft exosuits have potential to solve several design and control challenges, including achieving a high strength-to-weight ratio, inherent safety, comfort, and low cost, as well as avoiding joint misalignment between the body and the wearable device~\cite{perez2021soft}.

\begin{figure}[!t]
      \centering
      \includegraphics[width=\columnwidth]{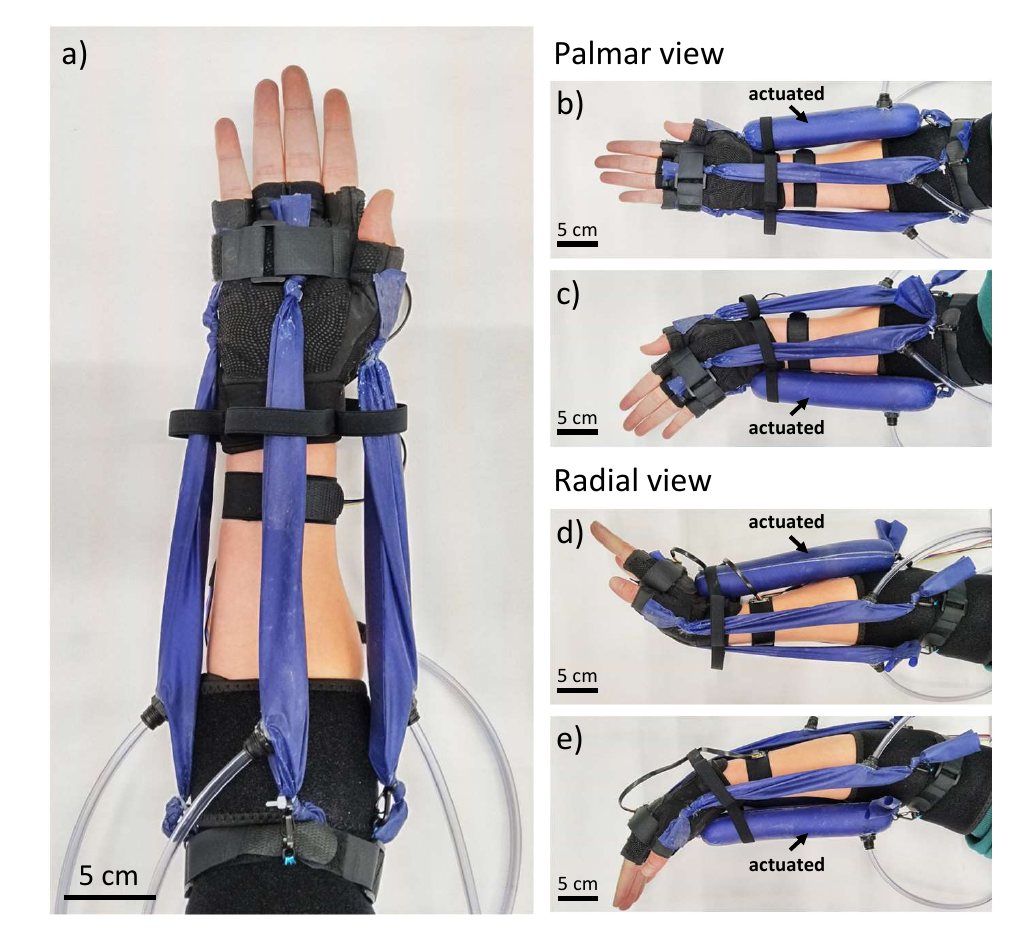}
      \vspace{-0.8 cm}
      \caption{Overview of our soft wrist exosuit. (a) Palmar view of the wrist exosuit with four fabric pneumatic artificial muscles (fPAMs)  (blue) uninflated. (b-c) Palmar view showing that the fPAMs at the two sides of the hand promote radial (top) and ulnar (bottom) deviation. (d-e) Radial view showing that the fPAM at the dorsal side promotes wrist extension (top) and at the palmar side promotes wrist flexion (bottom).}
      \label{fig:Overview}
      \vspace{-0.6 cm}
  \end{figure}

The advancements in soft robotics, specifically soft actuators~\cite{el2020softactuators, walker2020softpneumaticactuators}, have had a significant influence on the increase in the number of different exosuit designs. However, there are recently developed actuators whose application for exosuits remains unexplored. Our research is focused on examining the feasibility of one of these actuators, the fabric pneumatic artificial muscle (fPAM)~\cite{NaclerioRAL2020}, for movement assistance through a wrist exosuit (Fig.~\ref{fig:Overview}).

There are multiple actuation methods used for soft exosuits, each with their own advantages and limitations~\cite{thalman2020review, xiloyannis2021soft, perez2021soft}. The most commonly applied method is motorized, cable-driven actuation~\cite{thalman2020review}. Cable-driven systems have the benefits of producing high torques at a high bandwidth, being low-profile and having good controllability. McKibben type pneumatic artificial muscles, on the other hand, have higher power density~\cite{xiloyannis2021soft} and simple power transmission~\cite{perez2021soft} compared to cable-driven actuation. Nevertheless, McKibben actuators have multiple limitations as well, such as their bulkiness and slower response time compared to cable-driven systems, and their nonlinear behavior which makes their control challenging. Fabric pneumatic artificial muscles are similar to McKibben actuators, however, they are made of a single layer of bias-cut fabric and have advantages over McKibben muscles in various aspects. fPAMs are lightweight and fully foldable, and they have a near-linear force-contraction relationship, an absence of hysteresis, a high fatigue life, and a quick response to dynamic inputs~\cite{NaclerioRAL2020}. Exploring the advantages and limitations of this actuator for exosuits allows us to determine the optimal scenarios for its implementation.

The wrist joint serves as an advantageous foundation for upper limb exosuit design. The human wrist can move in two degrees of freedom: flexion/extension (F/E), and ulnar/radial deviation (U/R), with a potential third degree of freedom of forearm pronation/supination (P/S) depending on how close to the elbow an exosuit may be attached~\cite{palmer1985functional}. Devices developed for the wrist could potentially be adapted for other joints of the upper limb as needed by the user.

Rigid wrist exoskeletons~\cite{pezent2017design, zhang2020design} can provide high torque and wide range of motion for rehabilitation in a clinical setting, however various soft exosuits have been developed in recent years focusing on low cost and portability. While not strictly soft, the wrist exoskeletons presented in~\cite{higuma2017low} and~\cite{andrikopoulos2015motion} have a freely moving wearable part and overcome joint misalignment compared to completely rigid devices, but they lack other benefits of a fully soft design, such as garment-based anchoring and low profile wearable components. Other soft wrist exosuits apply cable-driven~\cite{choi2019exo, li2020bioinspired, chiaradia2021assistive}, shape memory alloy-based~\cite{jeong2019design}, McKibben pneumatic artificial muscle-based~\cite{bartlett2015soft, al2016wrist, andrikopoulos2015motion}, bending elastomeric~\cite{ang2019design, zhu2017carpal}, and textile pneumatic actuator-based~\cite{park2019lightweight, realmuto2019robotic} actuation.

\begin{table*}
\caption{Wrist Exoskeletons and Exosuits Using Linear Pneumatic Actuation\label{tab:wrist_exosuits}}
\vspace{-0.2 cm}
\label{specs}
\centering
\begin{tabular}{|>{\centering\arraybackslash}p{1.6cm}|>{\centering\arraybackslash}p{1.0cm}|>{\centering\arraybackslash}p{5.2cm}|>{\centering\arraybackslash}p{1.7cm}|>{\centering\arraybackslash}p{2.7cm}|>{\centering\arraybackslash}p{0.7cm}|>{\centering\arraybackslash}p{2.1cm}|}

\hline
\textbf{Author} & \textbf{DoF} & \textbf{Actuation (and Topology)}& \textbf{RoM} & \textbf{Torque} & \textbf{Weight} & \textbf{Control}\\
\hline
\hline
Zhang et al., 2019 \cite{zhang2020design} & 2 (F/E, U/R) & fixed exoskeleton with 2 pneumatic cylinders (they run parallel on ulnar side) & (65$^\circ$F, 65$^\circ$E), (80$^\circ$U, 25$^\circ$R) & 2.8 Nm cont. torque in F/E, 2.6 Nm in U/R (with 400 kPa) & - & unspecified motion control\\
\hline
Andrikopoulos and Manesis, 2015 \cite{andrikopoulos2015motion} & 2 (F/E, U/R) & portable exoskeleton with 4 Pneumatic Muscle Actuators (2 run parallel on dorsal, and 2 on palmar side) & (42$^\circ$F, 41$^\circ$E), (31$^\circ$U, 33$^\circ$R) & - & 430 g & advanced nonlinear PID-based control \\
\hline
Bartlett et al., 2015 \cite{bartlett2015soft} & 3 (F/E, U/R, P/S)  & exosuit with 4 McKibbens (2 are crossed on dorsal, and 2 are crossed on palmar side) & 91$^\circ$ in F/E, 32$^\circ$ in U/R, 78$^\circ$ in P/S & - & 220 g & - \\
\hline
Al-Fahaam et al., 2016 \cite{al2016wrist} & 2 (F/E, U/R) & exosuit with 5 McKibbens (3 contracting and 2 bending actuators on dorsal side) & - & - & 150 g & - \\
\hline
Park et al., 2019 \cite{park2019lightweight} & 1 (P/S) & exosuit with 2 flat pneumatic actuators (wrapped around the forearm symmetrically) & 180$^\circ$ in P/S & 0.78 Nm peak torque (with 50 kPa) & - & closed-loop angle control\\
\hline
Our design & 2 (F/E, U/R) & exosuit with 4 fPAMs (two pairs of antagonistic muscles on dorsal/palmar and radial/ulnar side) & (45$^\circ$F, 39$^\circ$E), (27$^\circ$U, 16$^\circ$R) & 3.3 Nm peak flexion torque (with 103 kPa) & 160 g & antagonistic feedback control\\
\hline
\end{tabular}\\
\vspace{-0.5 cm}
\end{table*}

Table~\ref{tab:wrist_exosuits} shows the specifications of wrist exoskeletons and exosuits that use linear pneumatic actuators, similarly to our exosuit, listing the degrees of freedom (DoF), actuation method and topology, range of motion (RoM), peak torque with corresponding pressure, weight, and implemented control if reported. The order of the exosuits follows increased softness. Our fPAM-actuated wrist exosuit stands out as it is thoroughly evaluated (which also includes the control), strong, lightweight, and, among the exosuits with fabric-based actuators, it is the first to actuate flexion/extension and ulnar/radial deviation.

In this paper, we present a novel soft wrist exosuit actuated by fabric pneumatic artificial muscles in antagonistic configuration (Fig.~\ref{fig:Overview}). First, we describe the exosuit prototype. Then, we introduce a two-dimensional geometric model of the exosuit to estimate the torque that the device applies to the human wrist. Next, we demonstrate the use of the model to select placement parameters of the fPAM mounting points, with the objective of closely approximating the peak biological flexion torques exhibited by the human wrist. Among the other pneumatic wrist exosuits~\cite{al2016wrist, andrikopoulos2015motion, ang2019design, bartlett2015soft, park2019lightweight, realmuto2019robotic, zhu2017carpal}, our model is the first to be applied for optimizing the exosuit design. We present the optimization results for an example user. We also experimentally validate the modeled torques throughout the range of wrist flexion/extension, which shows the significance of the stretching of fabric elements at the mounting point. Additionally, we compare the biological and exosuit-assisted range of motion in both flexion/extension and ulnar/radial deviation. Finally, we present a demonstration of the device actively controlled to move the wrist to a desired flexion/extension angle with a pair of antagonistic artificial muscles and to follow a desired two-degree-of-freedom trajectory by coordinating the movement of both pairs of actuators.

In summary, the main contributions of our work are:
\begin{itemize}
  \item application of the fPAM actuator (previously used to actuate robots~\cite{NaclerioRAL2020}) for movement assistance of the wrist;
  \item presenting and validating a torque model for the exosuit and introducing a stretching model to estimate mounting point displacement;
  \item design optimization to reach a desired torque with this actuator in antagonistic arrangement based on a simplified geometric model and using standard optimization tools;
  \item evaluation of the fPAM-actuated exosuit including the torque and range of motion measurement;
  \item demonstration of the fPAM-actuated exosuit through two trajectory tracking tasks on the passive human wrist with standard feedback control.
\end{itemize}

\vspace{-0.3 cm}
\section{Design and Fabrication} \label{sec:design}

This section discusses the considerations which led to the current design of the fPAM-actuated exosuit. After an overview of the design concept, we introduce the exosuit prototype in detail.

\vspace{-0.2 cm}
\subsection{Design Overview}

The main goal of our study is to utilize the advantageous properties of fPAMs for actuating soft exosuits. While there are several criteria for designing wearable devices, such as safety, ergonomics, autonomy, and cost \cite{perez2021soft}, here, we focus on creating the largest possible range of wrist motion and torque application. Because of the properties of the actuator, two other design criteria, safety and low cost of the wearable parts, are automatically satisfied.

Our wrist exosuit design (Fig.~\ref{fig:Overview}) consists of four pneumatic artificial muscles arranged radially around the wrist at 90$^\circ$ to each other. When each muscle is contracted upon applying positive pressure, it promotes one of the following wrist movements: flexion, extension, ulnar deviation, and radial deviation of the wrist. Due to the described placement of the fPAMs, there are antagonistic pairs of actuators in the sagittal and frontal planes. This actuator configuration allows movement limits to be an inherent part of the physical system. When one muscle inflates, the initially minimal resting force of the opposite side muscle increases while the muscle is stretched. The antagonist muscle acts as a hard stop when it reaches its maximum length change. Also, the limited contraction ratio of the agonist fPAM stops the movement at a given joint angle.

\begin{figure}[tb]
      \centering
      \includegraphics[width=\columnwidth]{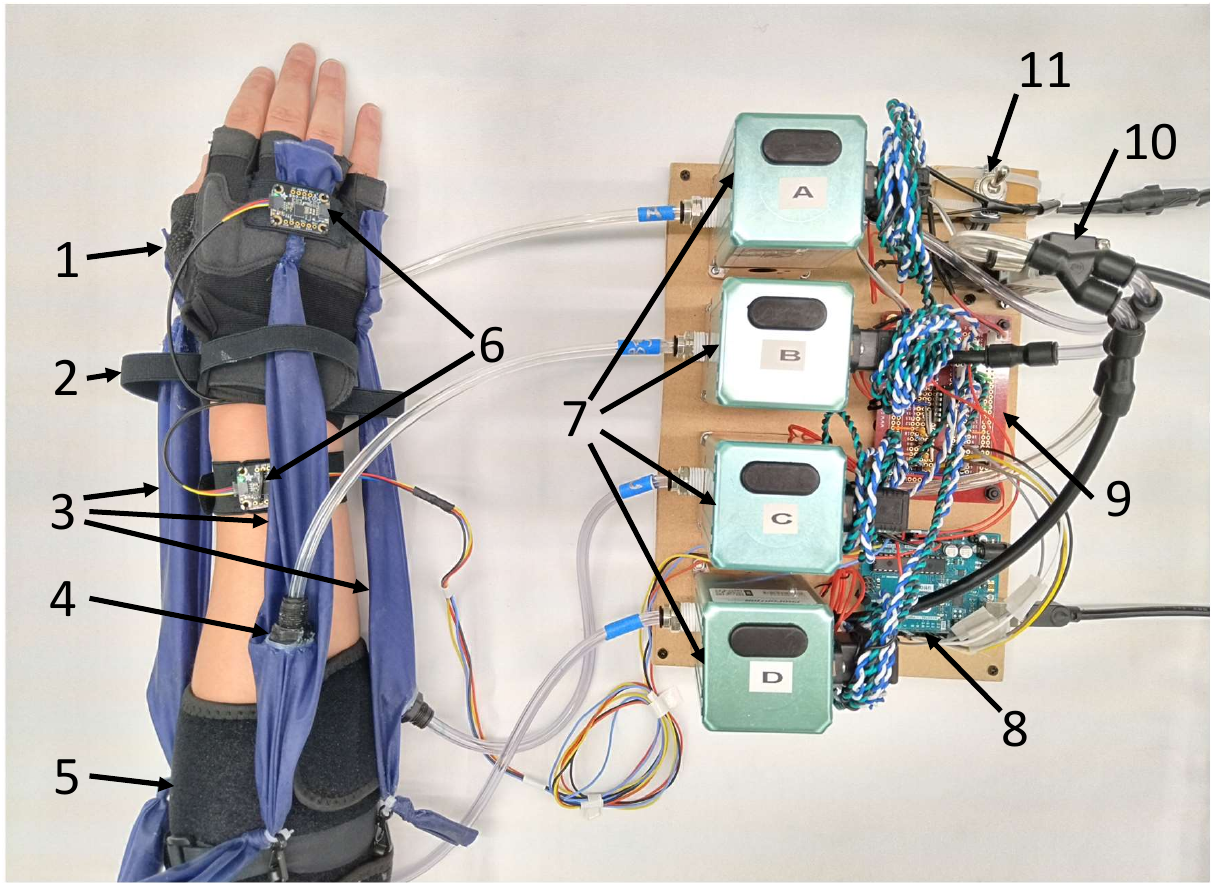}
      \vspace{-0.6 cm}
      \caption{Prototype of the wrist exosuit. The wearable components include the glove (1) with restricting bands (2), the four fPAMs (3) with push-to-connect pneumatic fittings (4), the elbow band (5) and two IMUs (6). The off-board elements include the pressure regulators (7), the microcontroller (8), the signal conditioning circuit (9), the solenoid valve (10), and a safety switch (11).}
      \label{fig:DesignDetail}
      \vspace{-0.3 cm}
  \end{figure}

\subsection{Exosuit Prototype}
\begin{figure}[tb]
      \centering
      \includegraphics[width=8.3 cm]{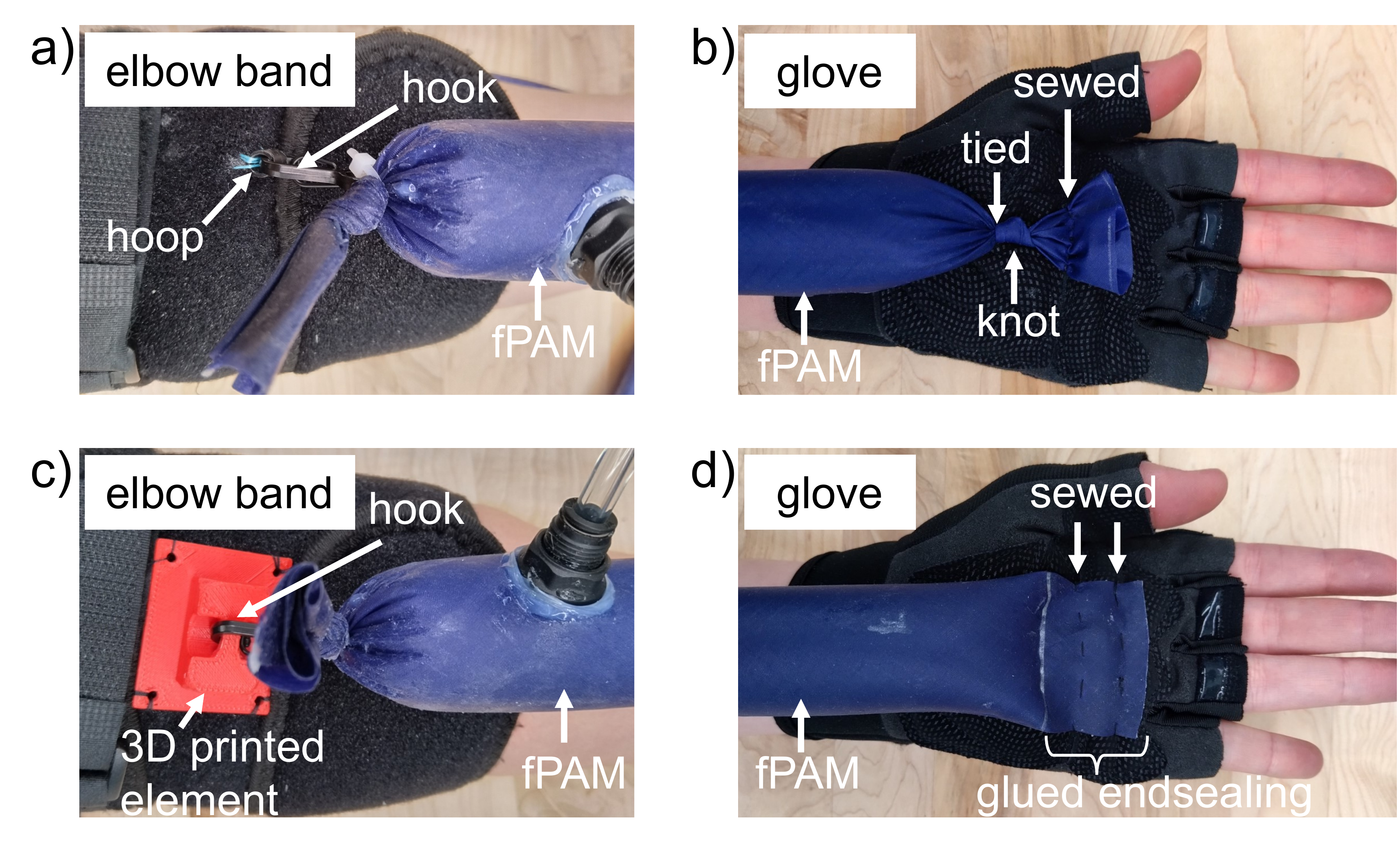}

      \caption{Various methods to attach the ends of the fPAM to the exosuit. The first row (a-b) shows the methods that were used for the final exosuit design and the second row (c-d) demonstrates the alternatives for attachment.}
      \label{fig:EndpointAttachment}
    \vspace{-0.4 cm}      
  \end{figure}

The exosuit prototype is shown in Figs.~\ref{fig:DesignDetail}~and~\ref{fig:EndpointAttachment}. The main wearable components are a fabric glove (component 1 in Fig.~\ref{fig:DesignDetail}), the fPAMs (component 3), a fabric elbow brace (component 5), and two inertial measurement units (IMUs) (BNO055, Adafruit) (component 6) having a total weight of approximately 160~g. One IMU is placed on the dorsal side of the hand, and another is attached to the dorsal side of the forearm in the same orientation when the wrist is straight (Fig.~\ref{fig:DesignDetail}). The relative orientation of the IMU sensors provides two-degree-of-freedom wrist angle information for the system.

We made the fPAMs out of silicone-coated ripstop nylon fabric (30 Denier Double Wall Ripstop Nylon Silicone Coated Both Sides, Rockywoods) based on the fabrication steps described in~\cite{NaclerioRAL2020} with the ends of the muscles sealed by tying a knot and using glue (Sil-Poxy, Smooth-On) to prevent its sliding. We added push-to-connect pneumatic fittings (component 4) reinforced with glue on the sides of muscles to attach the air tubes. One end of each fPAM is connected to a metal hook and attached to a ring sewed to the elbow brace (Fig.~\ref{fig:EndpointAttachment}(a)), while the other end is sewed to the glove (Fig.~\ref{fig:EndpointAttachment}(b)). This enables the muscles to be stretched tighter than their deflated resting length by attaching them to the elbow band after the glove and the band are put on. Each fPAM is routed through a restricting elastic band which is attached to the glove to ensure that the fPAMs can not slide off the surface of the wrist when they wrap around it.

An alternative method to connect the end of the fPAM to the elbow band is shown in Fig.~\ref{fig:EndpointAttachment}(c). In this case, a 3D-printed plastic piece is sewed to the band and serves as the structure to which the metal hook can be attached. Incorporating this rigid structure allows us to place the attachment point further away from the body but adds more non-fabric components. An alternative method for attaching the muscle to the glove is to seal the end of the fPAM by gluing the fabric together and then sewing it to the glove (Fig.~\ref{fig:EndpointAttachment}(d)). This method increases the low profile nature of the design, but it was not used here, as it proved to be less durable, and the position of the attachment point was more difficult to define.

Besides the wearable part, the exosuit consists of off-board components. The fPAMs are connected to closed-loop pressure regulators (QB3, Proportion-Air) (component 7 in Fig.~\ref{fig:DesignDetail}) through air tubes with a 6.35~mm outer diameter. The regulators can maintain the pressure level between 0 and 137~kPa depending on the input voltage. In the current configuration, a pressurized air source of 200~kPa connected to the regulators through a solenoid valve is used to supply air to the system. An Arduino Uno control board (component 8) is used to set the desired pressure level based on the angle of the wrist obtained by the IMUs using the control scheme described in Section~\ref{sec:control}. The control signal from the microcontroller goes through a signal conditioning circuit (component 9) consisting of a low-pass filter and a buffer to set the input voltage to the pressure regulators. The board is supplied by a 15~V DC power source, and it is equipped with a safety switch (component 11) that controls the solenoid valve (component 10) to cut off the air supplied to the system.

\vspace{-0.2 cm}
\section{Modeling} \label{sec:modeling}

In this section, we introduce a planar geometric model of our exosuit that will be used to calculate the torque applied by an fPAM to the wrist. We propose two torque equations, one describing the torque when the fPAM is running in a straight line between the mounting points, and a second describing the torque when the fPAM partially wraps around the wrist (e.g., the flexor fPAM in the fully extended wrist position).

\subsection{fPAM force modeling} \label{subsec:fPAM_modeling}

The force that an fPAM can apply at a given level of contraction can be calculated from~\cite{NaclerioRAL2020} and is described by Eqn.~\ref{eq:force}.
\begin{equation}
\label{eq:force}
\begin{aligned}
F=&\pi P(\frac{1}{\sin(\alpha_0)^2}-\frac{3(\epsilon-1)^2}{\tan(\alpha_0)^2})r_0^2 + 2 \pi E t (\epsilon_0-\epsilon)r_0\\
\text{where }\\
\epsilon =& (L_0 -L)/L_0 \\
\end{aligned}
\end{equation}
The produced force ($F$) of an fPAM with a given length ($L$) depends on the internal pressure ($P$), the contraction ratio ($\epsilon$) defined as the change in length over the original, fully stretched length of the fPAM $(L_0)$, the fully stretched fiber orientation of the fabric ($\alpha_0$), the fully stretched radius of the fPAM ($r_0$), the fabric thickness ($t$), and the elastic modulus of the fabric ($E$).
The equation is composed of two components added together. The first, pressure-dependent component is the ideal McKibben muscle model \cite{tondu2012modelling}, which represents the contraction force of a pressurized cylinder that reduces its length while its radius increases with the restriction of an unstretchable outer mesh. The second, pressure-independent component models the force that the stretched fabric applies as linear elasticity. The $\epsilon$ intercept point is denoted by $\epsilon_0$, which corresponds to the contraction ratio where the deflated fPAM starts to apply elastic force when it is stretched. We assign zero elastic force for contraction ratios larger than $\epsilon_0$.

The fPAM reaches its fully contracted length when the applied force becomes zero. Our tensile testing measurements (Section~\ref{subsec:measurement_of_the_fPAM_force}) show that the maximum contraction ratio ($\epsilon_{max}$) varies with pressure, therefore it is important to incorporate this variable into the force equation. To achieve this, we calculate the initial fiber orientation for each pressure by Eqn.~\ref{eq:fiber_orientation} \cite{NaclerioRAL2020}, so that the force equation implicitly includes $\epsilon_{max}$.

\begin{equation}
\label{eq:fiber_orientation}
\alpha_0 = -\arcsin{(\frac{\sqrt{\epsilon_{max}^2-2\epsilon_{max}+\frac{2}{3}}}{\epsilon_{max}-1})}
\end{equation}
This formula is derived by substituting $F=0$ and $\epsilon=\epsilon_{max}$ into Eqn.~\ref{eq:force}. At this value of $\epsilon$, the elastic, pressure-independent component is zero, and the equation can be rearranged to solve for $\alpha_0$.

\subsection{Exosuit torque model} \label{subsec:torue_modeling}

\begin{figure}[tb]
      \centering
      \includegraphics[width=8.0 cm]{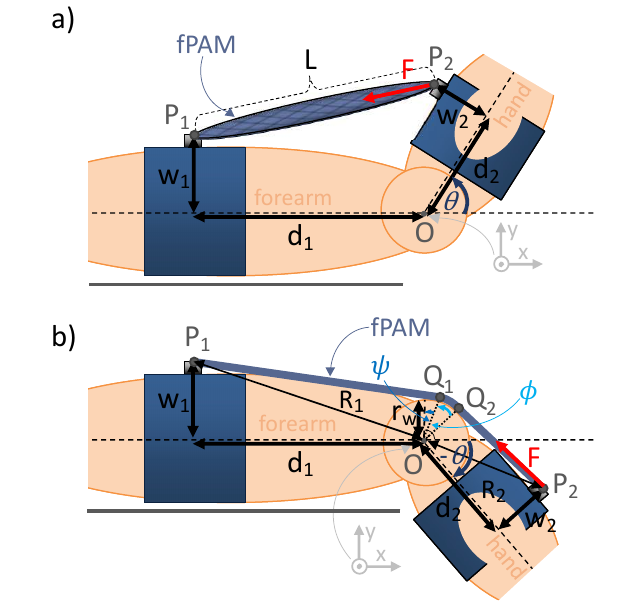}
      \vspace{-0.4 cm}
      \caption{2D geometric models of the torque applied by an fPAM to the wrist. (a) When the actuator does not wrap around the wrist, the fPAM runs in a straight line (e.g., the configuration of the fPAM on the palmar side of the hand at wrist flexion). (b) When the wrist is rotated in the opposite direction, the actuator wraps around the wrist.
      Model parameters and introduced notations are indicated in the figure.}
      \label{fig:GeometricModel}
      \vspace{-0.6 cm}
  \end{figure}

To approximate how much torque a single fPAM can apply to the wrist, we need to know the position of its two endpoints, relative to the center of rotation of the wrist. We used the two-dimensional geometric model (similar to the model used for the cable-driven exosuit in \cite{chiaradia2021assistive}) shown in Fig.~\ref{fig:GeometricModel} to calculate the torque that an actuator producing a given force ($F$) can apply to the wrist at a given wrist angle $\theta$. We use slightly different models for the torque computation at wrist angles where the actuator does not wrap around the wrist (Fig.~\ref{fig:GeometricModel} (a)) and where it does wrap around the wrist (Fig.~\ref{fig:GeometricModel} (b)). In both models, we make the following assumptions regarding the wrist kinematics and fPAM placement. The forearm and the hand are modeled as rigid links connected by a revolute joint. The two mounting points and the center lines of the two links are in the plane perpendicular to the axis of rotation. The axis intersects the plane at point $O$, which is defined as the center of rotation. In this plane, $d_1$ represents the projected length of the vector from the mounting point on the forearm ($P_1$) to the center of rotation ($O$) onto the center line of the forearm, and $w_1$ represents the shortest distance between the center line of the forearm and the mounting site on the forearm. The $d_2$ and $w_2$ distances denote the analogous placement parameters for the mounting point on the hand ($P_2$). The joint angle $\theta$ describing the orientation of the hand relative to the forearm is drawn in the positive direction on Fig.~\ref{fig:GeometricModel}(a) given that the reference coordinate frame is defined with the $z$ axis aligned with the axis of rotation and the $x$ axis parallel to the forearm, pointing in the direction of the hand. The figure illustrates flexion/extension movement, however, this model can be generalized to describe wrist rotation around another rotational axis of the wrist.

\subsubsection{Straight line model} \label{subsec:torue_modeling_straightline}
In the case where the actuator does not wrap around the wrist (Fig.~\ref{fig:GeometricModel}(a)), the modeled torque is given by Eqn.~\ref{eq:tau1} (similar to the torque equation of the cable-driven exosuit in \cite{chiaradia2021assistive}). This torque is calculated by the cross product between the vector from $O$ to $P_2$ and the vector along the fPAM in the direction from $P_2$ to $P_1$ with magnitude $F$. $L$ is the current length of the actuator, which is calculated as the Euclidean distance between the mounting points $P_1$ and $P_2$.
\begin{equation}
\label{eq:tau1}
\begin{aligned}
\tau =& F\frac{(d_1 d_2 - w_1 w_2) \sin(\theta) + (d_1 w_2 + d_2 w_1) \cos(\theta)}{L}\\
\text{where}\\
L = &\sqrt{
\begin{aligned}
(-&d_1-d_2 \cos(\theta) + w_2 \sin(\theta))^2 + \\
(&w_1 -d_2 \sin(\theta) -w_2 \cos(\theta))^2)
\end{aligned}
} 
\end{aligned}
\end{equation}

\subsubsection{Partially wrapped model} \label{subsec:torue_modeling_wrapping}
In the case when the actuator partially wraps around the wrist (Fig.~\ref{fig:GeometricModel}(b)), an extended geometric model with an approximated radius of curvature of the wrist can be used to calculate the torque. The wrist is assumed to have a circular surface with a constant radius ($r_w$), which defines the point where the fPAM first touches the wrist ($Q_1$) and the length of the segment where the fPAM is in contact with the wrist (from $Q_1$ to the last point of wrapping ($Q_2$)). The angle $\psi$ is defined as the angle between the $y$ axis and the vector from $O$ to $Q_1$, and the angle $\phi$ corresponds to the arc segment where the wrapping occurs (i.e., between $O$$Q_1$ and $O$$Q_2$). The $\psi$ angle is calculated by finding the position of $Q_1$ using Thales's theorem to construct the tangent from $P_1$ to the wrist circle. Also, the $R_1$ and $R_2$ parameters are assigned to denote the distances between the mounting points and the center of rotation. This allows us to define a closed-form expression for $\phi$ as given in Eqn.~\ref{eq:parameters2}.

\begin{equation}
\label{eq:parameters2}
  \begin{aligned}
   \phi = & \frac{\pi}{2} -\theta -(\psi+\arccos{(\frac{r_w}{R_2})}+\arcsin{(\frac{w_2}{R_2})}) \\
   \text{where}\\
    \psi=& -\arctan2((\hat{y} \times \vec{OQ_1}) \cdot \hat{z},\hat{y} \cdot \vec{OQ_1})\\
    \end{aligned}
\end{equation}
To calculate the torque, first, we need to consider the change in the length of the fPAM. We formulated a new equation (Eqn.~\ref{eq:length}), where the overall length of the muscle ($L_w$) during wrist wrapping is the sum of the distance between $P_1$ and $Q_1$, the arc length between the wrapping points $Q_1$ and $Q_2$ and the distance between $Q_2$ and $P_2$. The derived equation for the muscle length will be used to calculate the ideal fully stretched length of each fPAM by substituting the maximum wrist angle in the direction that causes wrapping into the equation.

\begin{equation}
\label{eq:length}
    L_w = \sqrt{R_1^2-r_w^2} + r_w\phi + \sqrt{R_2^2-r_w^2}
\end{equation}

Once we know the length of the fPAM, the magnitude of the fPAM force ($F$) at a given pressure can be computed using Eqn.~\ref{eq:force}. Compared to calculating the Euclidean distance between the endpoints, as in the case when the actuator does not wrap around the wrist, the length is increased, and therefore the magnitude of the force will be larger. The force will point in the direction from $P_2$ to $Q_2$. Similarly to the previous torque equation, the torque ($\tau_w$) during wrist wrapping is computed as the cross product of the vector from $O$ to $P_2$ and the force vector (Eqn.~\ref{eq:tau2}).

\begin{equation}
\label{eq:tau2}
\begin{aligned}
&\tau_w = \frac{Fr_w}{||P_2 Q_2||}((d_2\cos(\theta) - w_2\sin(\theta)) \cos(\phi-\psi) -\\
&\quad \quad \quad \quad  \quad \quad \;(d_2\sin(\theta) + w_2\cos(\theta) ) \sin(\phi-\psi)) \\
&\text{where}\\
&||P_2 Q_2|| = \sqrt{
\begin{aligned}
&(r_w\sin(\phi-\psi) -d_2\cos(\theta)+w_2\sin(\theta))^2 +\\
&(r_w\cos(\phi-\psi) -d_2\sin(\theta)-w_2\cos(\theta))^2
\end{aligned}
}
\end{aligned}
\end{equation}

The transition between the straight-line and wrapping model occurs when the fPAM touches the surface of the wrist while the joint extends. For the 2D geometric model, this occurs when the line along the two mounting points changes from being disjoint to becoming tangent to the circle representing the wrist. To decide which model to use for a given joint angle, we examine the number of intersection points between the line along $P_1$ and $P_2$ (assuming the general form of $y=ax+b$) and the wrist circle ($x^2+y^2={r_w}^2$) by solving the system of equations for the $x$ and $y$ coordinates of the shared points. This leads to a quadratic equation in either $x$ or $y$, therefore the discriminant of the quadratic formula ($D$) (Eqn.~\ref{eq:straight_to_wrapping}) can be used to determine the number of solutions.
  \begin{equation}
   \label{eq:straight_to_wrapping}
     \begin{aligned}
    & D={2ab}^2-4(1+a^2)(b^2-{r_w}^2)\\
      \end{aligned}
  \end{equation}
The transition from the straight line to the wrapping model happens when there is one intersection point ($D=0$).

\section{Design Optimization} \label{sec:design_optimization}

One of the key challenges of this exosuit design is to find the best placement for the actuator mounting points. There are advanced optimization methods for cable-driven upper limb exosuits that utilize biomechanical models (e.g., musculoskeletal models in OpenSim~\cite{delp2007opensim}). This approach has proven to be beneficial, for example, to minimize muscle activation for complex joint structures like the shoulder~\cite{joshi2022designoptimization_shoulder} or to minimize required cable tension in the case of a less complex joint such as the elbow over the routing geometries of multiple cables~\cite{bardi2022designoptimization_elbow}. Given the lower complexity of our exosuit design and the wrist joint, we propose a simplified, kinematic model-based methodology for the selection of parameters governing the design of fPAMs and their positioning on the body.

In this section, we demonstrate how the torque model can be used to find these parameters such that the exosuit can apply as high torque as the maximum biological torque in any wrist configuration. To solve this task, we present a parameter optimization methodology using standard optimization tools
to fit the modeled torque to the reference biological data and conduct an example optimization for a single, healthy user.

\subsection{Reference biological torque} \label{subsec:biol_torque}

Based on the force-contraction ratio relationship of the fPAM \cite{NaclerioRAL2020}, this actuator can provide high forces when the contraction ratio is close to zero, but the force approximately linearly decreases to zero as the muscle contracts. Therefore, we expect that it is more difficult for the exosuit to match the biological torques when the torque values follow a uniform distribution over a large joint angle range. We chose the flexion torque over the flexion/extension range to be the reference data, as this torque profile is the closest to the previously described criteria, and thus it is the most challenging for the exosuit to reach.

\begin{figure}[tb]
      \centering
      \includegraphics[width=8.3cm]{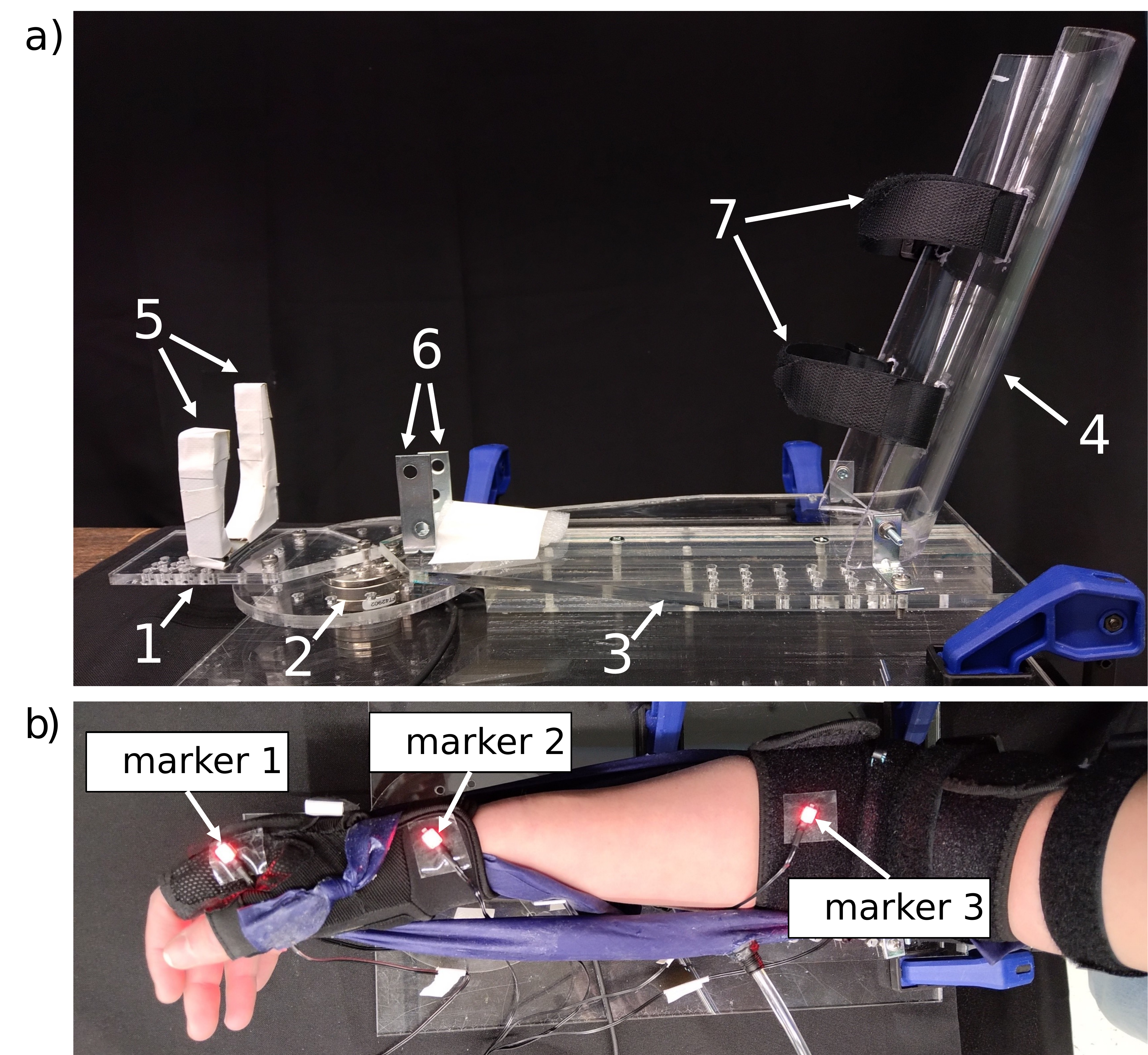}
      \vspace{-0.3 cm}
      \caption{Measurement setup for the peak biological wrist torque and the applied exosuit torque over wrist flexion/extension. (a) Side view of the torque measurement setup (without the human arm). The main components include the hand plate with adjustable angle (1), the torque sensor (2), the static structural elements (3,4), and additional support structures (5,6,7) to fix the arm. (b) Placement of the three motion capture markers used to measure the wrist angle.}  \label{fig:BiologicalTorqueMeasurement}
      \vspace{-0.3 cm}
  \end{figure}

To collect peak biological wrist torque data, we conducted a torque measurement on a single subject using the setup in Fig.~\ref{fig:BiologicalTorqueMeasurement}, which is made of acrylic sheet and tube elements. The hand plate (component 1 in Fig.~\ref{fig:BiologicalTorqueMeasurement}(a)) is connected to the torque sensor (Mini 45, ATI) (component 2), which has a circular plate on the top side and which is fixed to the base plate on the bottom side. The hand plate is directly attached to the circular plate such that the orientation of the hand plate can be changed using screws to fix the angle of the wrist in a desired position between 90$^\circ$ extension and 90$^\circ$ flexion with an increment of 22.5$^\circ$. The forearm plate (component 3) reaches over the circular plate to provide support for the forearm, but it is only attached to the base plate. At the other end of the forearm plate, a tilted half cylinder (component 4) connects to it to secure the position of the upper arm. Additional support structures, i.e., brackets and hook-and-loop tapes (components 5, 6, and 7), prevent the horizontal displacement of the hand, forearm, and upper arm.

Throughout the torque measurement, the wrist angle was monitored by recording the position of three motion capture markers (Impulse X2E, PhaseSpace) on the exosuit. The locations of the markers were defined based on the marker placement in \cite{schmidt1999marker} adding some changes due to the measurement setup limiting the available surfaces on the radial side of the arm. The marker at the hand is on the head of the second metacarpal, the marker at the wrist is on the styloid process of the radius, and the marker at the elbow is approximately halfway between the medial and lateral epicondyle of the humerus. Fig.~\ref{fig:BiologicalTorqueMeasurement}(b) illustrates the markers placed on the hand (marker~1), the wrist (marker~2), and the forearm (marker~3). Because the vertical positions of the markers are not the same, the vectors between the markers were projected into the horizontal plane when the angle between them was calculated.

The peak voluntary flexion torque was measured three times (while the fPAMs were not actuated) varying the angle of the hand plate from 67.5$^\circ$ extension to 90$^\circ$ flexion in steps of 22.5$^\circ$. The average biological peak torque with standard deviation is shown in Fig.~\ref{fig:ParameterOptimizationResult}, along with the parameter optimization results described in the next subsection. It is important to note that the measured reference data is not generalized for multiple users, however, the results provide the torque profile of a healthy individual capable of performing activities of daily living. Moving forward, the findings of the paper discuss the results of this case study.

\subsection{Parameter optimization} \label{subsec:param_opt}

\begin{figure}[tb]
      \centering
      \includegraphics[width=\columnwidth]{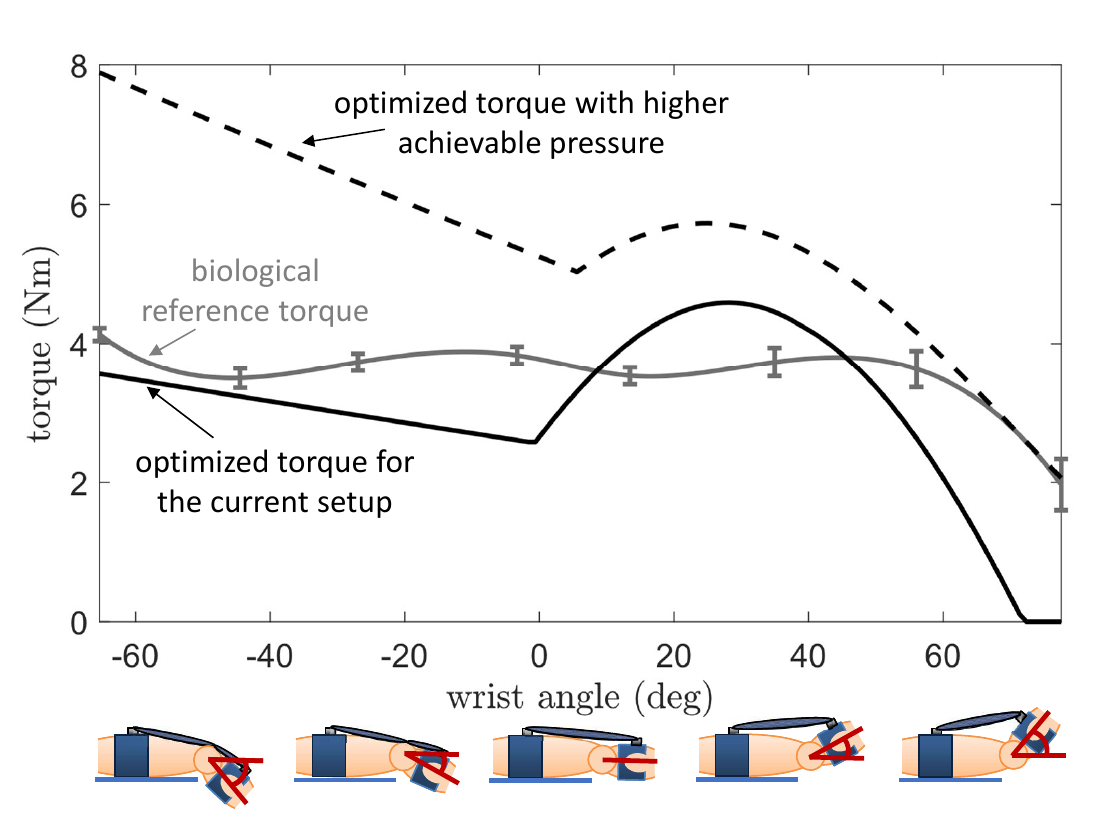}
      \vspace{-0.8 cm}
      \caption{Results of the parameter optimization for exosuit flexion torque across extension (negative) and flexion (positive) angles. The grey curve corresponds to the average measured (standard deviation indicated with error bars) and interpolated peak biological wrist flexion torques of a single user. The continuous black curve shows the predicted exosuit torque for our current setup with the optimized parameters shown in Table~\ref{tab:optimized_parameters}. The dashed black curve shows the result of the same optimization process for an alternative physical system that can provide higher pressure, 312~kPa instead of our current 137~kPa upper limit.}
      \label{fig:ParameterOptimizationResult}
      \vspace{-0.3 cm}
  \end{figure}

The goal of the parameter optimization is to find the parameters that allow the exosuit to achieve at least as high torque as the biological reference torque ($\tau_{ref}$) at all wrist angles. The collected average biological peak torque data described in the previous section was used for the optimization, with the number of data samples ($N$) increased from 8 to 140 using ``spline" interpolation in MATLAB (continuous grey line in Fig.~\ref{fig:ParameterOptimizationResult}). This made the optimization more independent of the specific angles chosen for the measurement points.
We formulated an objective function ($f$) given in Eqn.~\ref{eq:objective_function} as the sum of the positive differences between the biological reference torque ($\tau_{ref}$) and the modeled exosuit torque ($\tau_{mod}$) across the full range of joint angles ($\theta_i, i=1,...,N$).
  \begin{equation}
   \label{eq:objective_function}
     \begin{aligned}
    & f(X)=\sum_{i=1}^{N} \textrm{max}(0,\tau_{ref@\theta_i}-\tau_{mod}(X,\theta_i))\\
      \end{aligned}
  \end{equation}

 When minimizing this objective function over a set of exosuit parameters ($X$), a mismatch between the model and reference data is penalized when the modeled exosuit torque is lower than the reference torque.

The exosuit parameters can be divided into two groups. The first group includes the fPAM parameters, which describe the actuator itself as in the modeling equations: the fully stretched radius ($r_0$) and length ($L_0$), and the internal pressure ($P$). The second group includes the placement parameters, which describe where the fPAMs are attached to the glove and elbow band ($d_1$, $w_1$, $d_2$, and $w_2$ in Fig.~\ref{fig:GeometricModel}).

In the first group, the fully stretched radius and the initial length depend on the fabrication process of the fPAM, and the applied pressure can be changed through the operation of the exosuit. Because the torque is proportional to the magnitude of the fPAM force, the force should be maximized for the purpose of optimization. The pressure and the square of the radius are both proportional to the magnitude of the ideal, pressure-dependent force, and the radius is proportional to the elastic force (Eqn.~\ref{eq:force}), so these parameters should be fixed at their highest value. The highest achievable pressure based on our current physical system is 137~kPa, which corresponds to the maximum output pressure of the pressure regulators. For this optimization, we consider the maximal contraction ratio ($\epsilon_{max}$) to be 0.28, as that corresponds to 137~kPa in~\cite{NaclerioRAL2020}. We chose to set the fully stretched radius to 1.23~cm, which corresponds to the widest fPAM (with approximately 5~cm diameter when unstretched and uninflated) that we fabricated and subjectively considered to be low profile in the exosuit. For different design considerations and system properties (e.g., different pressure regulators), the choice of these parameter values is restricted only by the inverse proportional relationship between the maximum pressure and the fully stretched radius~\cite{NaclerioRAL2020}. Similarly, the force magnitude is higher if the muscle is less contracted, therefore the initial, fully stretched length of the fPAM should be the shortest length that still allows the user to reach the full range of motion in the direction that stretches the muscle. We computed the initial length using Eqn.~\ref{eq:length} with the maximal extension angle to match these criteria. For this optimization, we defined the maximal extension angle as $-$65.4$^\circ$, which corresponds to the largest extension angle where we measured reference peak biological torque. Because we used one specific type of silicone-coated ripstop nylon fabric, the fabric thickness ($t$) and the elastic modulus of the fabric ($E$) are constant parameters. The reported thickness of the fabric by the manufacturer is 0.08~mm, and the value that we used for elastic modulus is 9.06~MPa based on our discussion with the authors of \cite{NaclerioRAL2020}.

The placement parameters determine the moment arm and the required actuator length change, as well as the direction of the actuator force, which changes throughout the range of motion. Therefore, it is not straightforward how they influence the exosuit torque. We conducted a design optimization on the four placement parameters by minimizing the objective function (Eqn.~\ref{eq:objective_function}). We defined upper and lower bounds on each parameter (Table ~\ref{tab:optimized_parameters}) based on the dimensions of the arm of the current user and our capability to make a firm attachment point on the glove and the elbow brace. For $d_1$ and $d_2$, the lower bounds ensure that we can still place a short fPAM across the wrist by keeping a small distance from the center of rotation both on the forearm and the hand. The upper bounds are based on the fact that the lengths of the hand and forearm limit how far away the fPAM can be placed from the wrist. For $w_1$ and $w_2$, the lower bounds are based on the smallest width measured on the hand and forearm, and the upper bounds are based on increasing these distances by adding an elevating structural component to the exosuit (Fig.~\ref{fig:EndpointAttachment}(c)).
\begin{table}[tb]
\caption{The bounds on each parameter, and the results of the parameter optimization\label{tab:optimized_parameters}}
\vspace{-0.2 cm}
\centering
\begin{tabular}{||p{4.5 cm} p{0.5 cm} p{0.5 cm} p{0.5 cm} p{0.5 cm}||}  
     \hline
                     & $d_1$ & $w_1$ & $d_2$ & $w_2$\\ 
     \hline\hline
     Lower bound [cm]   & 2.00   & 3.50  & 1.50 & 1.50\\ 
     \hline
     Upper bound [cm]   & 22.00 &   5.00 & 9.00 & 3.50\\
     \hline
     Resolution for initial guesses [cm]  & 2.00 & 0.50 & 0.50  & 0.50\\
     \hline
     Optimized value [cm] & 22.00 & 3.50 & 9.00  & 1.50\\
     \hline
    \hline
\end{tabular}
\vspace{-0.4 cm}
\end{table}

The last parameter that appears in the torque equations is the radius of the wrist. This parameter is independent of the exosuit configuration. The radius was computed as half of the measured width of the human wrist in the sagittal plane, which is 2~cm for the user in this study.

For the optimization, we used the modified version of the built-in function \textit{fminsearch} in the MATLAB Optimization Toolbox that allows the definition of upper and lower bounds of the parameters. The running time of a single cycle of the algorithm is on average 0.9 seconds. Because it is not guaranteed to converge to the global minimum, we ran the optimization with a set of initial guesses. This set contained points of the discretized parameter space with the resolution presented in Table~\ref{tab:optimized_parameters}. The running time of the full optimization process was 24 minutes on average.
 
 The torque with the optimized parameter values is plotted in Fig.~\ref{fig:ParameterOptimizationResult} across the flexion/extension wrist angles. The optimization results show that the optimized exosuit torque (black curve in Fig.~\ref{fig:ParameterOptimizationResult}) is smaller than the biological peak torque (grey curve) for all wrist extension angles, with a value close to the reference torque at full extension but linearly decreasing until reaching $-$1$^\circ$. However, the torque increases at the early stage of wrist flexion and becomes higher than the reference torque for a set of flexion angles starting from 9$^\circ$. The exosuit torque becomes lower than the biological peak torque at 46$^\circ$ and continues decreasing until it reaches zero at 72$^\circ$. When the wrist moves from full extension to $-$1$^\circ$ of extension (close to the neutral position), the model describing the fPAM wrapping around the wrist defines the torque (Eqn.~\ref{eq:tau2}). In this range, the moment arm is constant, but the force and thus the torque decreases as the muscle length shortens compared to the fully stretched length. When the wrist is rotated from $-$1$^\circ$ of extension towards full flexion (with torque described by Eqn.~\ref{eq:tau1}), the muscle runs in a straight line between the mounting points, and its length monotonically decreases. Although initially the torque increases due to the sudden increase in the moment arm, the torque eventually approaches zero as the moment arm gradually stops rising and the fPAM continues to produce smaller forces. This result indicates that, given the set of constraints we have placed on our design, an fPAM-based exosuit for this specific user cannot reach the same peak torque over the whole range of movement as the human wrist. Note, however, that the typical functional range of human wrist motion (35$^\circ$ of flexion/extension ~\cite{palmer1985functional}) is smaller than the full joint range, and on this smaller range, the exosuit should reach close to the peak biological torque.

 Besides the parameter optimization for our current setup, we ran a search to find the smallest pressure level that provides enough torque to exceed the reference torque for all joint angles. We found that the required pressure is 312~kPa, which is below 448~kPa, the burst pressure of the fPAM with the implemented fully stretched radius of 1.23~cm, calculated as described in~\cite{NaclerioRAL2020}. The optimal placement parameters remained the same compared to the previous optimization, except $d_2$ was decreased to 3.14~cm. This indicates that, based on the model prediction, if the pressure regulators of the physical system are changed to reach this required pressure, the exosuit torque (dashed black curve in Fig.~\ref{fig:ParameterOptimizationResult}) should exceed the biological torque over the full flexion/extension range.

\section{Model Validation and Experimental Results} \label{sec:model_validation_and_experimental_results}

In this section, we describe the procedure and results of the measurement of the torque of the physical exosuit with only the flexor muscle attached, as well as the process of identifying model parameters. We propose adjustments to the model that increase its accuracy by modeling the displacement of the mounting point due to fabric stretching. We also present the results of a measurement to compare the biological and exosuit-actuated range of motion of the human wrist.

\subsection{Measurement of the fPAM force} \label{subsec:measurement_of_the_fPAM_force}

\begin{figure}[tb]
      \centering
      \includegraphics[width=\columnwidth]{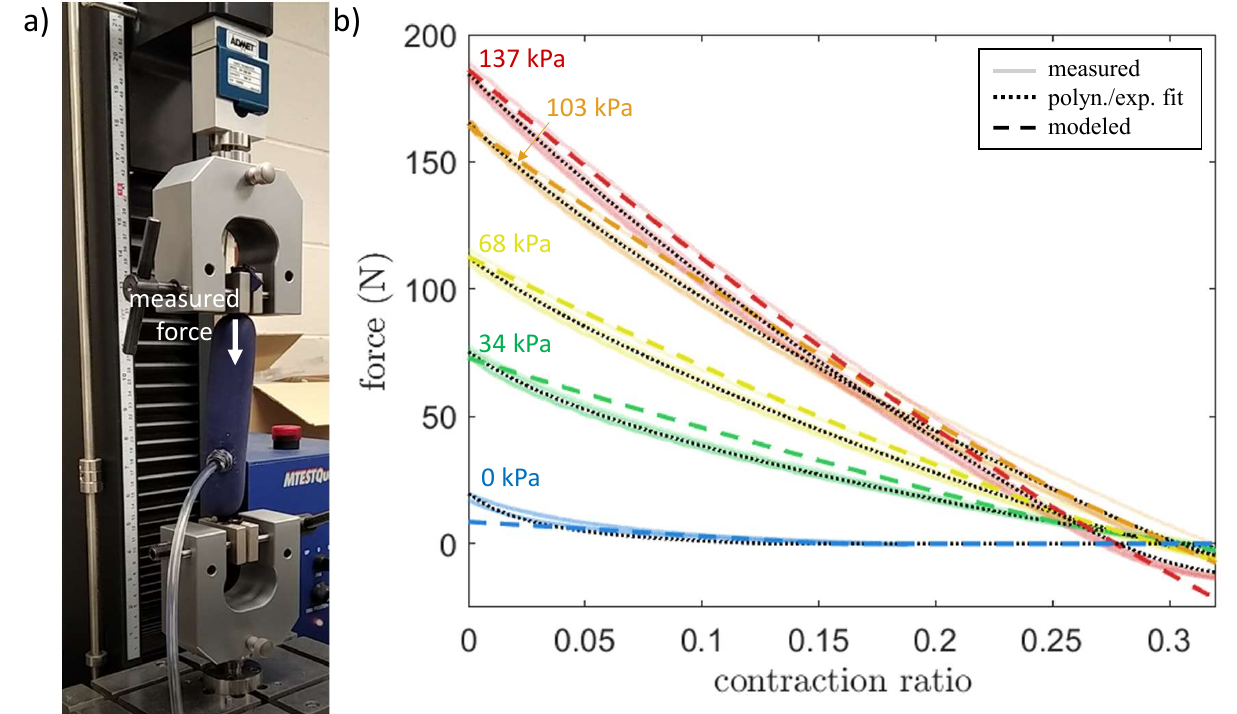}
      \vspace{-0.6 cm}
      \caption{Tensile testing measurement of the flexor fPAM used in the exosuit torque validation of Fig.~\ref{fig:ExperimentalData}. (a) Tensile testing setup. (b) Measured fPAM force-contraction profile of the flexor muscle. The solid lines are the measured forces at pressures of 0 kPa (blue), 34 kPa (green), 68 kPa (yellow), 103 kPa (orange), and 137 kPa (red). The individual measurements are plotted with faded colors.
      The polynomial and, for the zero pressure curve, the exponential approximations are plotted with black dotted lines. The colored dashed lines represent the fitted modeled force for each pressure level using Eqn.~\ref{eq:force} and the adjusted parameters in Table~\ref{tab:tensile_testing_parameters}.}
      \label{fig:TensileTest}
      \vspace{-0.3 cm}
\end{figure}

Before attaching the flexor fPAM on the exosuit, we conducted a tensile testing measurement (Fig.~\ref{fig:TensileTest}) to identify the parameters of the modeled force of the actuator. During the measurement process, the fPAM was stretched from its initial, fully contracted state (measured on the inflated fPAM at a given pressure when the force readings approached -10~N) to zero contraction (measured on the deflated fPAM when the applied load was 230~N as in~\cite{NaclerioRAL2020}) and then it was returned to its initial state. The linear force of the fPAM was measured by a load cell (SM-1000-961, ADMET) for three cycles of length change. The measurement was repeated for five pressure levels (0 kPa, 34 kPa, 68 kPa, 103 kPa, and 137 kPa). The test setup is shown in Fig.~\ref{fig:TensileTest}(a), and the measurement data is shown in Fig.~\ref{fig:TensileTest}(b).

For each nonzero pressure level, the smoothed and averaged force-contraction ratio plots were approximated by an 8th-order polynomial so that we could sample points at arbitrary contraction ratios. The same polynomial approximation does not work well for the zero-pressure elastic force, therefore a piecewise function consisting of an exponential and a constant zero function was used for approximating the fPAM force at zero pressure. 
The initial length $L_0$ and radius $r_0$ of the fully stretched but uninflated fPAM were measured, and the maximum contraction ratio was derived from the zero crossing of each force curve for each different pressure level. Still, an adjustment of $\epsilon_{max}$ and $r_0$ was required to match the modeled force to the measured data. Table~\ref{tab:tensile_testing_parameters} contains the derived fPAM parameters with $\epsilon_0$ denoted as the $\epsilon_{max}$ value at zero pressure. The angle of the fiber orientation $\alpha_0$ was calculated based on the derived maximum contraction ratio as described in the model of the fPAM (Eqn.~\ref{eq:fiber_orientation}). The modeled force based on Eqn.~\ref{eq:force} is plotted in Fig.~\ref{fig:TensileTest}(b) using the calculated parameter values.

  \begin{table}[!tb]
\caption{The fPAM model parameters for the fabricated flexor artificial muscle\label{tab:tensile_testing_parameters}}
\vspace{-0.2 cm}
\centering
    \begin{tabular}{||p{2.3 cm} p{0.7 cm} p{0.7 cm} p{0.7 cm} p{0.7 cm} p{0.7 cm}||}     
     \hline
      $P$ [kPa] & 0 & 34 & 68 & 103 & 137 \\
     \hline
     $L_0$ [cm] (measured)    & 34.0  & 34.0  & 34.0 & 34.0 & 34.0\\ 
     \hline
     $r_0$ [cm] (measured)   & 1.02  & 1.02  & 1.02 & 1.02 & 1.02\\
     \hline
     $r_0$ [cm] (calculated)    & 1.23  & 1.27  & 1.22 & 1.19 & 1.26\\  
     \hline
     $\epsilon_{max}$ (calculated)     & 0.153  & 0.303  & 0.296 & 0.301 & 0.277\\  
     \hline
\end{tabular}
\vspace{-0.4 cm}
\end{table}

\vspace{-0.2 cm}
\subsection{Torque applied by the exosuit} \label{subsec:torque_applied_by_the_exosuit}

If the model is validated for a single muscle without the other muscles being attached, we can confirm the applicability of the model for each fPAM due their similar placement. Then, the net torque can be approximated with the model as the sum of the torques of antagonistic fPAMs. Thus, we conducted the validation only for the flexor fPAM.

\begin{figure}[tb]
      \centering
      \includegraphics[width=8.3cm]{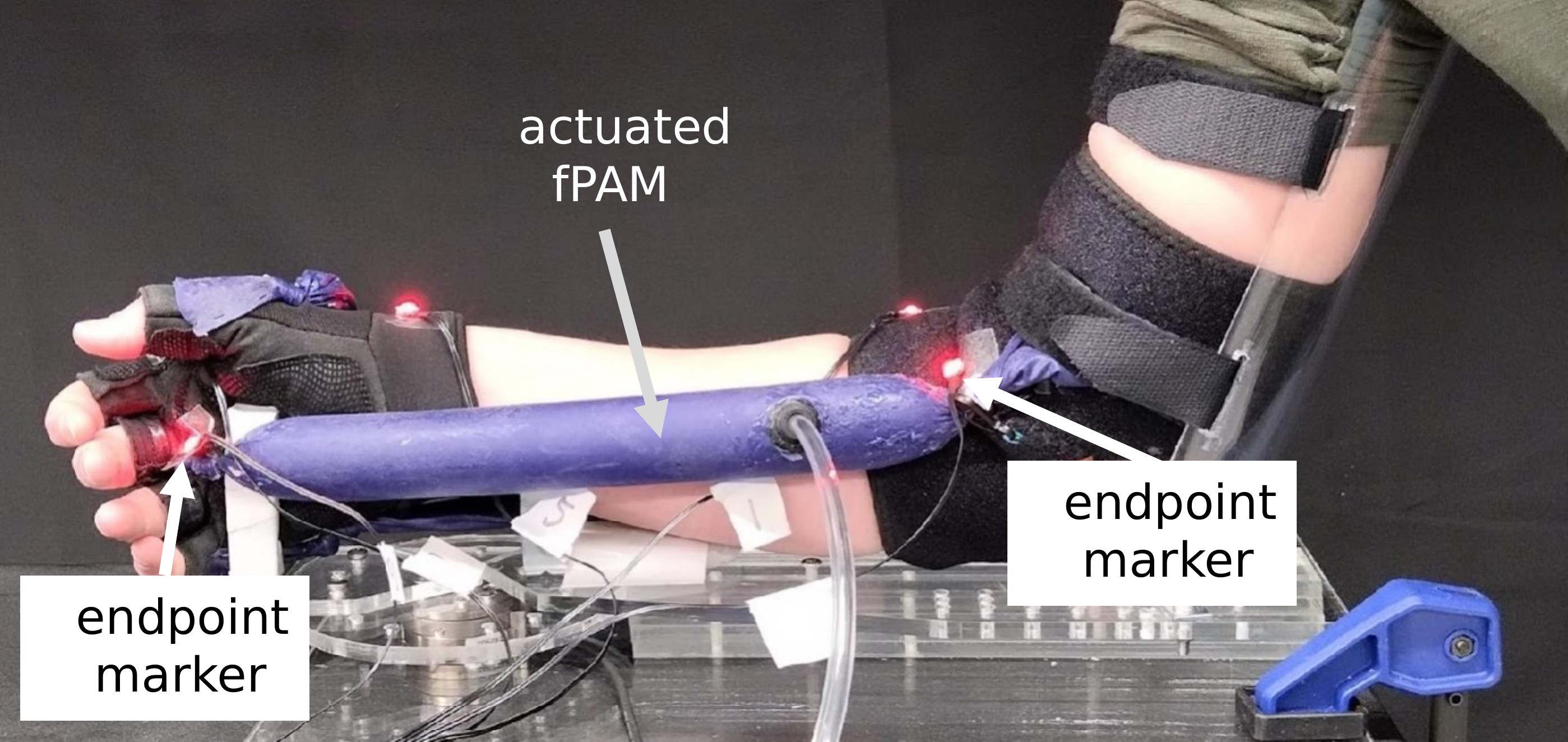}
      \vspace{-0.3 cm}
      \caption{Side view of the torque measurement setup which was previously presented in Fig.~\ref{fig:BiologicalTorqueMeasurement}. The current configuration shows how this experimental setup was used to measure the exosuit-applied flexion torque with the user's wrist relaxed at a fixed joint angle of nominally 0$^{\circ}$ (the forearm and hand plate are aligned). In addition to the wrist position markers (Fig.~\ref{fig:BiologicalTorqueMeasurement}(b)), motion capture markers were placed on the endpoints of the fPAM.}
      \label{fig:ExperimentalSetup}
      \vspace{-0.4 cm}
  \end{figure}

 We used a torque sensor (Mini 45, ATI) to measure the torques applied by the flexor muscle of the exosuit to the relaxed wrist over the range of wrist angles from $-$67.5$^\circ$ to 90$^\circ$ with an increment of 22.5$^\circ$ on the same experimental setup which was used for the measurement of biological peak torques previously presented (Fig.~\ref{fig:BiologicalTorqueMeasurement}). Similarly to that experiment, the measurement was repeated for five pressure levels (0 kPa, 34 kPa, 68 kPa, 103 kPa, and 137 kPa) three times at each wrist angle. The arm of the user was attached to the setup as illustrated in Fig.~\ref{fig:ExperimentalSetup}. The attachment point of the fPAM on the hand was placed close to the fingers to leave space for the brackets in the middle of the palm to brace the hand. The other mounting point was placed on the forearm as close to the elbow as possible.
 
 To track the positions of the endpoints relative to the human arm, motion capture markers were placed on the radial side of the exosuit to measure the current wrist angle (as before), and also on the palmar side of the exosuit right over the knot at the end of the fPAM as shown in Fig.~\ref{fig:ExperimentalSetup}. The placement parameters corresponding to the two-dimensional geometric wrist model were derived from the horizontal components of the marker coordinates. The parameters from Table~\ref{tab:tensile_testing_parameters} were used to model the fPAM force, but the initial length of the fPAM was reduced to 32.0~cm to make the actuator more stretched at full wrist flexion given the actual placement. The radius of the wrist ($r_w$) was determined by running an exhaustive search over a range of 1~cm to 7~cm with a resolution of 0.01~cm to minimize the sum of the RMS error between the measured and modeled torques for the two models over all data points.

\begin{figure}[tb]
      \centering
      \includegraphics[width=\columnwidth]{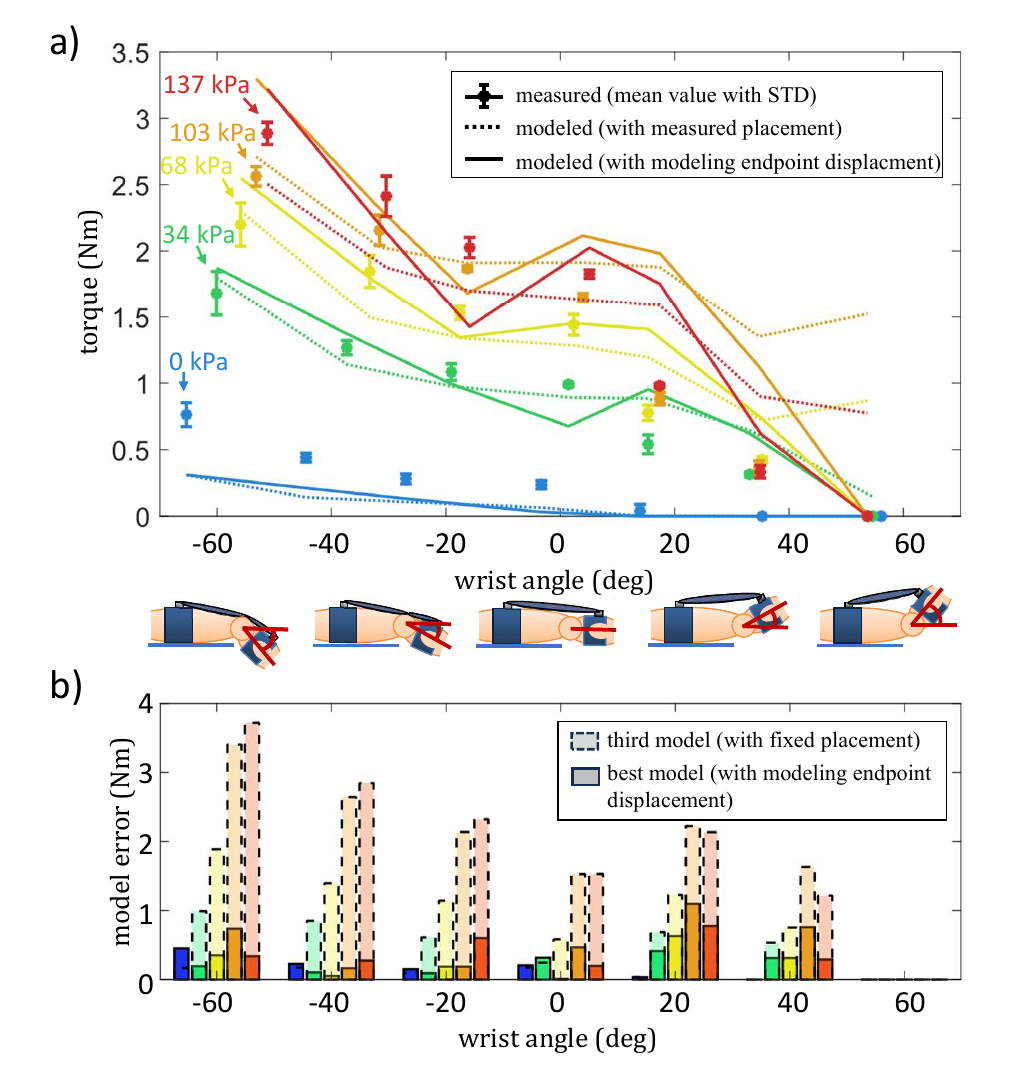}
      \vspace{-1.0 cm}
      \caption{Comparison between measured and modeled exosuit torques as a function of wrist angle. (a) Average and standard deviation of the three torque measurements is plotted for each wrist angle (dots). The measurement was repeated for five pressure levels. The torques were modeled in two ways: first, using the current position of the fPAM from motion capture data (dashed lines), and second, fixing the placement parameters and modeling the effect of the attachment point stretching (Eqn.~\ref{eq:fabric_stretching}) (solid lines). (b) Comparison between the average model error magnitude when the torque is modeled the second way (solid bars) and a third way (relying on fixed placement parameters and not modeling the endpoint stretching) (dashed bars).}
      \label{fig:ExperimentalData}
      \vspace{-0.3 cm}
  \end{figure}

Fig.~\ref{fig:ExperimentalData}(a) shows the measured and modeled torque values based on actuating the flexor fPAM over the defined flexion/extension range. The dots correspond to the average measured exosuit torques at each wrist angle, where the colors indicate the applied pressure going from 0 kPa (blue) to 137 kPa (red). 
We used two approaches to derive the placement parameters for the exosuit torque model. The first method (dashed lines) uses the actuator endpoint position data from the motion capture system at each measurement point. The second method (solid lines) uses fixed, initial placement parameters that correspond to the zero wrist position with the actuator deflated and extends the model to predict the mounting point displacement at each data point.

For the second method, we chose a spring equation with a second order displacement term (Eqn.~\ref{eq:fabric_stretching}) to describe the mounting point displacement, which represents the net effect of the fabric stretching of the elbow band and the glove, combined with the translation of the soft tissue of the arm. The second order displacement model was chosen, as it provided better fit to the measured data than a first order model.

\begin{equation}
    \label{eq:fabric_stretching}
    F=K_i{\Delta x_i}^2, \quad i\in \left\{ 1,2\right\}
\end{equation}
The indices differentiate the sites of the actuator mounting points, where $1$ corresponds to the forearm and $2$ corresponds to the hand. The force ($F$) equals the magnitude of the force applied by the fPAM at the endpoints using the initial placement parameters (Table~\ref{tab:torque_measurement_parameters}). The $\Delta x_i$ denotes the displacement of each endpoint in the direction of the force compared to the initial (unstretched) placement, and $K_i$ denotes the stretching coefficient.

\begin{table}[!t]
\caption{The parameters of the physical device during the torque measurements\label{tab:torque_measurement_parameters}}
\vspace{-0.2 cm}
\centering
    \begin{tabular}{||p{0.45 cm} p{0.45 cm} p{0.45 cm} p{0.45 cm} p{0.45 cm} p{0.45 cm} p{0.8 cm}  p{1.2 cm}||}  
     \hline
        $d_1$ [cm] & $w_1$ [cm] & $d_2$ [cm] & $w_2$ [cm] & $L_0$  [cm] & $r_w$  [cm] & $K_1$ [$N/{cm}^2$]& $K_2$ [$N/{cm}^2$]\\ 
     \hline\hline
        19.47  & 4.93  & 8.30 & 3.38 & 32.00 & 3.99 & $-$43.36 & $-$1097.50\\ 
     \hline
\end{tabular}
\vspace{-0.4 cm}
\end{table}

First, we derived the stretching coefficients by using the measured displacement, which was calculated based on the endpoint position data at each measurement point. Given the force and displacement, $K_i$ was computed at each measurement point by using Eqn.~\ref{eq:fabric_stretching}, and these values were averaged to derive a single value for the coefficient (Table~\ref{tab:torque_measurement_parameters}). When the coefficients are known, the torque can be computed only using the initial position data. Thus, to compute the torque, the endpoint displacement was calculated by using Eqn.~\ref{eq:fabric_stretching} with the derived $K_i$ values, and then the torque model was applied assuming that the fPAM endpoints deviate from the initial placement by this estimated displacement.
 
As shown in Fig.~\ref{fig:ExperimentalData}(b), adding the estimated displacement significantly improves the torque estimate compared to only relying on fixed parameters and ignoring the stretching (this latter referred to as the third method). The mean absolute error (MAE) of the stretching model based on the second method's predictions is 0.283~Nm, which is 27.7\% of the average magnitude of the measured torques, while the MAE of this third model is 1.111~Nm, which is 108.7\% of the average torque magnitude. Additionally, the second method approximates the torque also better than the first method, especially when the joint angle approaches the fully flexed position (shown in Fig.~\ref{fig:ExperimentalData}(a)). The MAE of the first method is 0.350~Nm, which is 34.3\% of the average torque magnitude.

\vspace{-0.2 cm}
\subsection{Range of motion} \label{subsec:range_of_motion}

\begin{figure*}[tb]
      \centering
      \includegraphics[width=\textwidth]{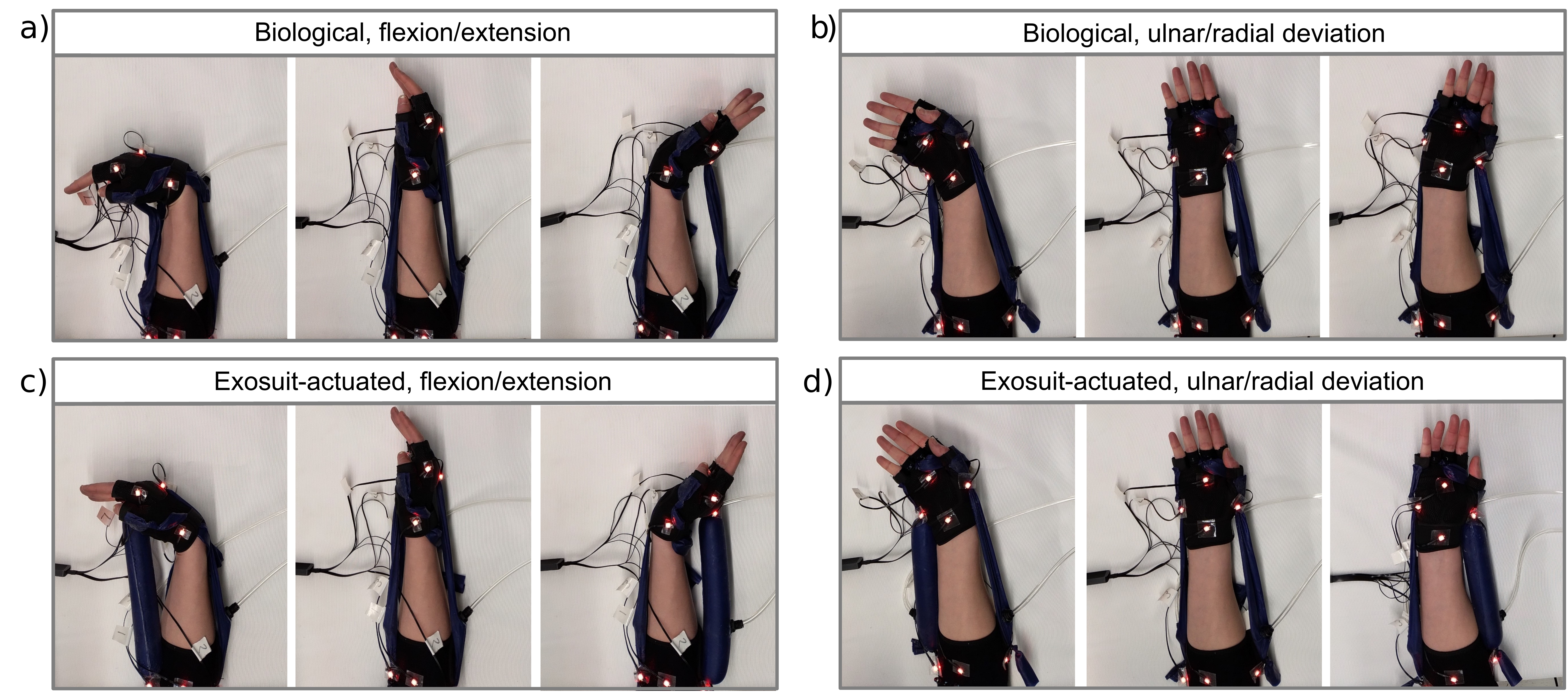}
      \vspace{-0.6 cm}
      \caption{Range of motion measurement with motion capture system. Three motion capture markers were placed on the mid-line of the forearm, center of the wrist, and mid-line of the hand to measure wrist angle. The endpoints of the actuated muscles were tracked by markers as well. (a) Fully flexed, neutral, and fully extended position of the voluntarily moved biological wrist when the exosuit is not actuated. (b) Full ulnar deviation, neutral, and full radial deviation position of the biological wrist. (c) Range of motion of the relaxed wrist in the flexion/extension plane is achieved by inflating the corresponding fPAMs of the exosuit (applying 137 kPa pressure). (d) Range of motion of the relaxed wrist in the ulnar/radial deviation plane when moved by the exosuit.}
      \label{fig:RoM}
      \vspace{-0.3 cm}
  \end{figure*}

We also conducted a measurement to compare the active biological and the exosuit-actuated range of motion of the human wrist. Motion capture markers were placed on the exosuit similarly as for the torque measurement (Fig.~\ref{fig:ExperimentalSetup}), but this time markers were placed on both the radial and the palmar side of the arm to collect angle data along two different planes. The endpoints of the actuated muscle were also tracked. 
The extensor muscle was attached to the hand just below the head of the third metacarpophalangeal joint. The radial deviation muscle was placed close to the base of the first metacarpal (the base of the thumb) and the ulnar deviation muscle was mounted symmetrically to the opposite side of the hand. The other mounting points for all fPAMs were placed on the forearm as close to the elbow as possible. The measurement process consisted of two parts. First, the exosuit-actuated wrist range was measured when the arm of the user was relaxed and the exosuit was actuated by applying 137 kPa pressure to the corresponding fPAM to passively move the human wrist to maximum flexion/extension (Fig.~\ref{fig:RoM}(a)) and ulnar/radial deviation (Fig.~\ref{fig:RoM}(b)) while the forearm was placed on the table. Then, the same measurement was repeated when the user of the exosuit voluntarily moved the wrist and the fPAMs on the exosuit were not actuated (Fig.~\ref{fig:RoM}(c) and (d)). Table~\ref{tab:RoM} contains the average and the standard deviation of three range of motion measurements for both the exosuit-actuated and the voluntarily moved wrist.

To derive the modeled range of motion, the same measurement was repeated once with tracking the endpoints of the actuated muscles at the joint limits. The placement parameters shown in Table~\ref{tab:RoM} were derived from the measured endpoint position at the joint limits using the same method as for the torque measurement. The torque was estimated from 0$^\circ$ to 90$^\circ$ along the four movement directions using the fixed placement parameters as we focusing on providing accurate prediction close to the joint limits. We assumed that the fPAM parameters $r_0$ and $\epsilon_{max}$ are the same as for the measured flexor fPAM at 137~kPa for all the muscles, and we measured the fully stretched length for all fPAMs separately. The modeled joint limits were defined as the wrist angle where the modeled torque reached zero. The model error was computed by subtracting the measured exosuit angle limits from the modeled angle limits. The modeled limits were higher in all cases, so the computed error values are positive (Table~\ref{tab:RoM}).

\begin{table}[!t]
\caption{Measured exosuit-actuated and biological range of motion of the wrist, and derived model parameters\label{tab:RoM}}
\vspace{-0.2 cm}
\centering
    \begin{tabular}{||p{1.5 cm} p{1.2 cm} p{1.2 cm} p{1.2 cm} p{1.2 cm} ||} 
     \hline
                &flexion & extension & ulnar dev. & radial dev. \\ 
     \hline\hline
        exosuit  & 44.5$^\circ$ (STD~7.5$^\circ$) & 38.7$^\circ$ (STD~5.2$^\circ$) & 26.8$^\circ$ (STD~3.4$^\circ$) & 15.9$^\circ$ (STD~3.3$^\circ$) \\
        \hline
        biological  & 80.9$^\circ$ (STD~6.7$^\circ$) & 56.0$^\circ$ (STD~6.8$^\circ$) & 39.1$^\circ$ (STD~1.3$^\circ$) & 23.2$^\circ$ (STD~2.3$^\circ$) \\
        \hline
        model error  & 5.3$^\circ$ & 20.9$^\circ$ & 16.1$^\circ$  & 14.0$^\circ$\\
        \hline
        $d_1$  & 20.5 cm & 24.5 cm & 21.1 cm  & 21.3 cm\\
        \hline
        $w_1$  & 5.7 cm & 3.7 cm & 7.1 cm  & 4.8 cm\\
        \hline
        $d_2$  & 4.0 cm & 5.4 cm & 1.0 cm  & 2.3 cm\\
        \hline
        $w_2$  & 4.0 cm & 4.1 cm & 3.4 cm  & 6.0 cm\\
        \hline
        $L_0$ (meas.)  & 29.0 cm & 32.0 cm & 28.5 cm  & 27.0 cm\\
    \hline
\end{tabular}
\vspace{-0.4 cm}
\end{table}
  
\section{Control and Demonstration} \label{sec:control}

\begin{figure}[tb]
    \centering
    \vspace{-0.2 cm}
    \includegraphics[width=\columnwidth]{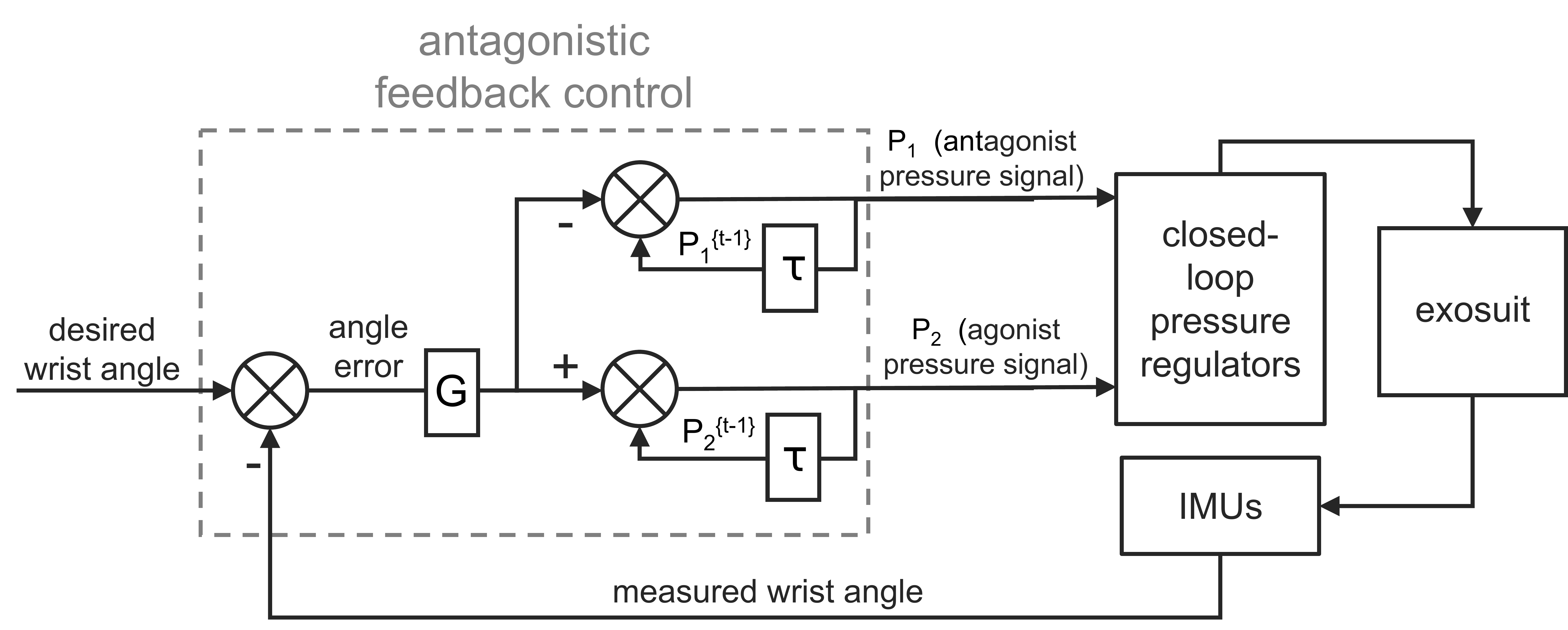}
    \caption{Diagram of the applied closed-loop control system. The desired trajectory is expressed in joint angles along a given degree of freedom. IMUs measure the actual wrist angle, and the derived angle error is used to update the pressure in the antagonistic pair of fPAMs. The error multiplied by a gain ($G$) is subtracted from the pressure in the antagonist artificial muscle at the previous iteration ($P_1^{\{t-1\}}$), and it is added to the pressure in the agonist muscle at the previous iteration ($P_2^{\{t-1\}}$).}
    \label{fig:Control}
    \vspace{-0.6 cm}
\end{figure}

To demonstrate how the prototype of the wrist exosuit functions to move the human wrist, we implemented a control algorithm specific to the antagonistic muscle configuration to automatically position the wrist at a desired joint configuration (Fig.~\ref{fig:Control}). We applied feedback control, as it is easy to implement and only requires measured actual wrist angle information from the two IMUs (measured at 100 Hz). Although this simple feedback control algorithm has limited precision, it was successfully used for the two trajectory tracking tasks described in this section to demonstrate the application of the exosuit on the human wrist.

\subsection{Control system} \label{subsec:control_system}

Based on the measured actual wrist angle and the desired wrist angle, the wrist angle error is used to change the pressure of the antagonistic muscles that control each physical angle. To produce antagonistic coordination, the error, scaled by a gain, is added to the current pressure of the agonist muscle that moves the wrist closer to the desired angle and subtracted from the current pressure of the antagonist muscle resisting the movement. We set the value of the gain to $0.0083$~kPa/deg (manually tuned based on observing the performance of the controller), which helps to convert the error in degrees to a small change of pressure in kPa at each time step through the control loop, which in our implementation had a time step of 0.014~s. Co-contraction of the two muscles causes increased stiffness. The initial stiffness depends on the initial pressures, which were set to 13.8 kPa for all muscles, but the stiffness was not directly regulated during operation. The stiffness, however, can increase during operation when, for example, the wrist movement is restricted and the pressure builds up in one muscle. To be able to decrease this built-up stiffness, we included an additional condition that the pressure of the resisting fPAM should decrease twice as fast if its current pressure exceeds the chosen threshold of 13.8~kPa.

\subsection{Planar angle tracking} \label{subsec:planar_angle_tracking}

As a first demonstration of exosuit control, we studied how our antagonistic feedback controller can position the relaxed human wrist along the flexion/extension direction in the horizontal plane by coordinating the operation of two artificial muscles. First, the wrist angle was set to zero when the wrist was in a neutral resting position. When the trajectory tracking started, the wrist was at an angle of 0$^\circ$ and the pressure in the muscles was set to 13.8 kPa. The desired wrist position was gradually increased by 10$^\circ$ to move the wrist towards the extension direction. The wrist angle remained constant for 10 seconds before moving to the next position. From 30$^\circ$ of wrist extension, the desired wrist angle was gradually decreased in the same manner to reach 40$^\circ$ of wrist flexion.

\begin{figure}[tb]
    \centering
    \includegraphics[width=\columnwidth]{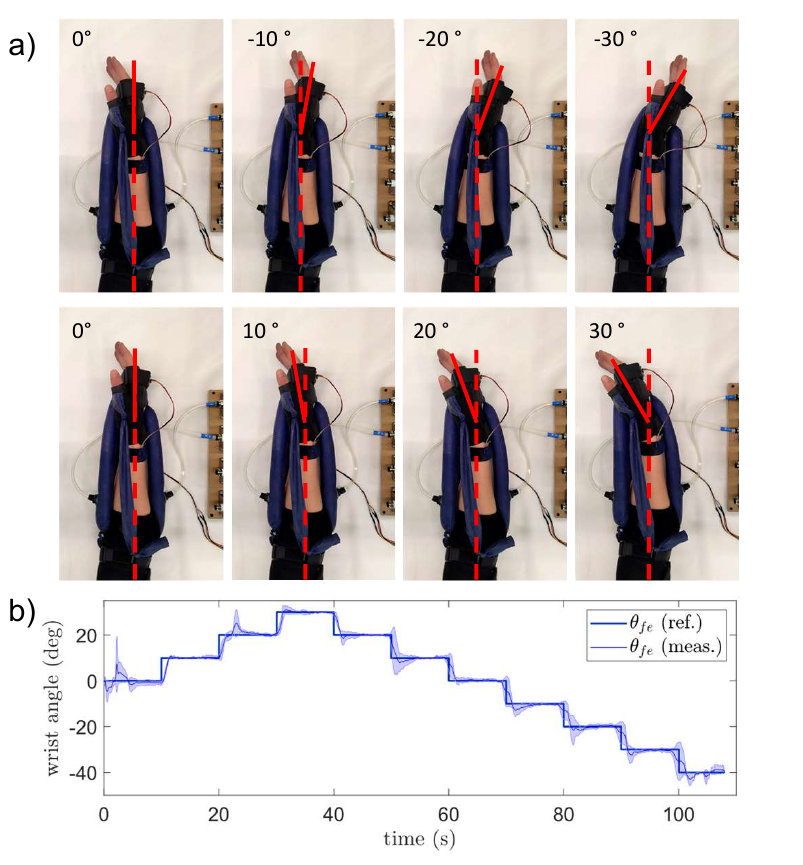}
    \vspace{-0.8 cm}
    \caption{Demonstration of the planar positioning of the wrist using antagonistic feedback control. (a) Series of pictures illustrating the actual wrist position when the desired wrist angle is increased from 0$^\circ$ to 30$^\circ$ of wrist extension (first row) and flexion (second row) by steps of 10$^\circ$. The nominal wrist angles are shown in red. (b) Plot showing the desired (thick line) and the average (thin lines) and standard deviation (shaded region) of three measured trajectories during the trajectory tracking task.}
    \label{fig:Demonstration_positioning}
    \vspace{-0.2 cm}
\end{figure}

Fig.~\ref{fig:Demonstration_positioning} illustrates the exosuit completing the planar positioning task by showing the exosuit at different goal positions (Fig.~\ref{fig:Demonstration_positioning}(a))
 and showing the desired trajectory and the measured trajectory during three trajectory tracking trials (Fig.~\ref{fig:Demonstration_positioning}(b)). The results confirm that the exosuit is able to reach the desired wrist angles, however, we can observe inaccuracies in the tracking when the desired position changes. The step responses of the three actual trajectories when increasing and decreasing the joint angle have an average of 0.84~s and 0.54~s rise time, and 19$\%$ and 44$\%$ overshoot, respectively.

\subsection{Spatial trajectory tracking} \label{subsec:spatial_trajectory_tracking}

As a second demonstration, we studied how our controller can control the wrist motion along its two degrees of freedom. We defined a sinusoidal desired trajectory in joint space for both flexion/extension ($\theta_{fe}$) and ulnar/radial deviation ($\theta_{ur}$) angles to imitate the tracing of a near-elliptic trajectory by the hand. The amplitudes of the sinusoidal trajectories were defined to be within the range of motion such that the flexion/extension angles are between $-$40$^\circ$ and 30$^\circ$ and the ulnar/radial deviation angles are between $-$10$^\circ$ and 30$^\circ$. Both trajectories have a period of 24 seconds, but there is a 90$^\circ$ phase shift between them. Similarly to the other trajectory tracking task, the joint angles were set to zero at a neutral position of the wrist before starting the tracking.

\begin{figure}[tb]
    \centering
    \includegraphics[width=\columnwidth]{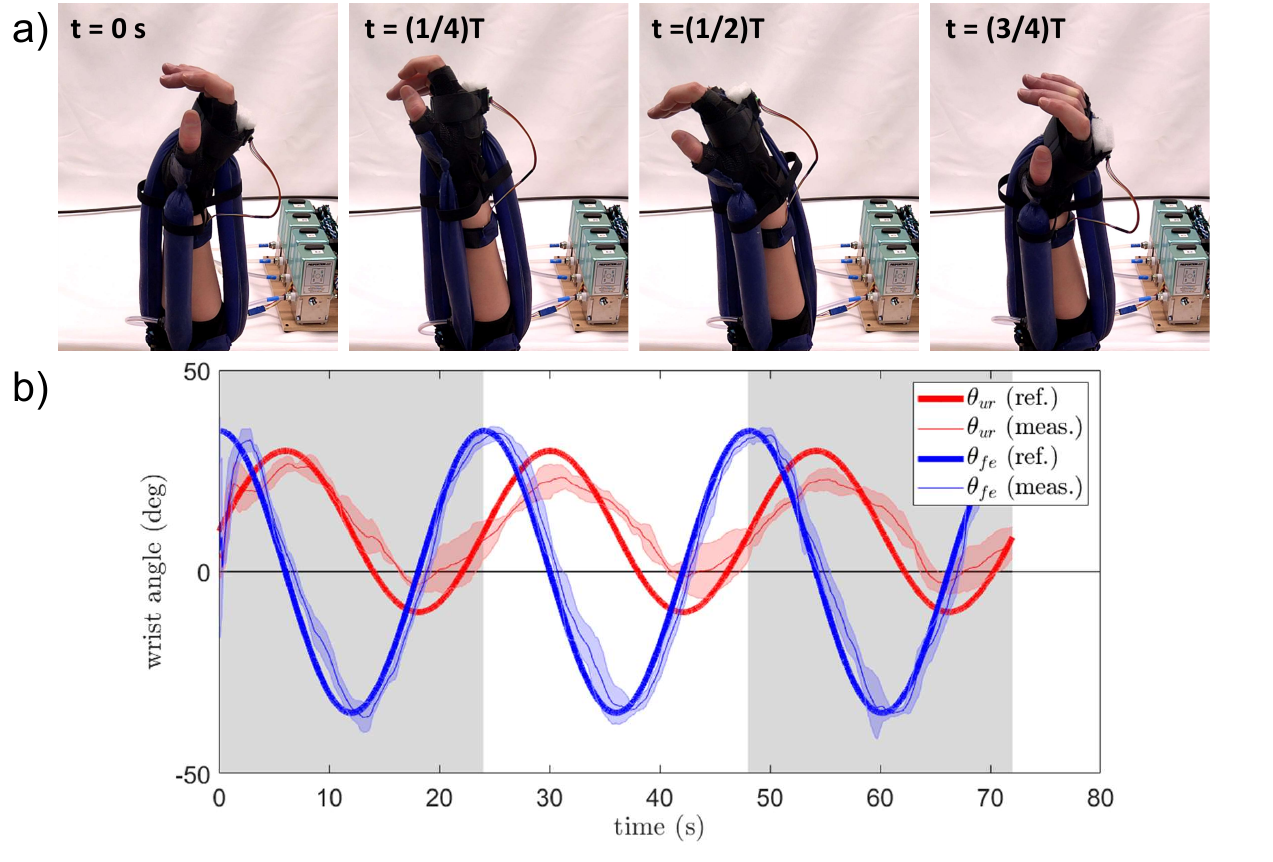}
    \vspace{-0.8 cm}
    \caption{Demonstration of spatial trajectory tracking by coordinating the actuation of all four muscles of the exosuit. (a) Series of pictures from the trajectory followed by the wrist. (b) Reference trajectory and average and standard deviation of four measured trajectories through the duration of three periods.}
    \label{fig:Demonstration_trajectory}
    \vspace{-0.5 cm}
\end{figure}

The results of the spatial trajectory tracking are shown in Fig.~\ref{fig:Demonstration_trajectory}. Fig.~\ref{fig:Demonstration_trajectory}(a) illustrates the realized motion of the wrist by showing the wrist position on four equally spaced points of the trajectory. Fig.~\ref{fig:Demonstration_trajectory}(b) shows the desired and four actual trajectories for the duration of three periods. The root mean square angle error of the tracking (not including the short initial settling phase) is 5.18$^\circ$ for flexion/extension and 7.12$^\circ$ for ulnar/radial deviation.

\vspace{-0.3 cm}
\section{Discussion} \label{sec:discussion}

In this section, we discuss the results of the presented work, starting with the parameter optimization and then focusing on the experimental results and demonstration. Finally, we present proposed movement assistance applications based on the observed properties of the exosuit.

\vspace{-0.1 cm}
\subsection{Parameter optimization}

\textbf{The modeled exosuit torques reach biological peak torques on a restricted range.} At the pressure allowed by the current setup, the optimized torques over the flexion/extension range are close to or higher than the biological reference torques, except for high wrist flexion angles (over 46$^\circ$ for the example user). This is due to the decreasing fPAM force towards full contraction.

\textbf{The optimized parameter values indicate that the optimization problem could be simplified.} The optimized placement parameters (Table \ref{tab:optimized_parameters}) were equal to the upper or lower bounds. Even when the fPAM pressure was increased, we reached the optimal torque profile when the mounting points were placed as close as possible to the skin (minimal $w_1$ and $w_2$) and as far as possible from the wrist on the forearm (maximal $d_1$). Although these optimization results could be different for other users with their specific arm dimensions and desired torque profiles, the current results highlight the importance of finding the right distance between the wrist and the mounting point on the hand ($d_2$).

\textbf{It is possible to reach higher torque with increased fPAM parameters.} The optimization with increased maximum pressure shows that the reference torque can be reached on a larger range due to higher torque. Therefore, it is advantageous to increase the internal pressure or the diameter of the fPAM. The increase of the diameter is only limited by the practical consideration of the increasing bulkiness as a larger fPAM is inflated. The input pressure is limited by the chosen pressure regulators or pressure source.

\subsection{Results of the torque measurement}

\textbf{It is important to correctly identify the fPAM parameters for accurate modeling.} The modeled force did not match the measured tensile testing data when we used the measured fully stretched radius ($r_0$) value, therefore this parameter had to be adjusted. Also, the force did not scale proportionally with the pressure due to the different maximum contraction ratio ($e_{max}$) values.
The tensile testing measurement proved to be a good method to calculate the parameters and, thus, to approximate the fPAM force with the model. However, it is important to conduct a more comprehensive analysis to understand the relationship between pressure and maximum contraction ratio.

\textbf{The modeling of the fabric stretching helps to tackle some of the challenges of identifying the fPAM placement parameters.} We examined different approaches to handle the displacement of the fPAM mounting points which occurs due to the stretching of the soft elements. When the torque model was used with fixed fPAM placement, the modeling error was significantly higher than the error with the modeling method that accounts for the displacement as shown in Fig.~\ref{fig:ExperimentalData}(b).

Also, two methods were introduced to account for endpoint displacement, which resulted in closer torque estimates. The first method directly used the actuator endpoint position data from the motion capture system. Compared to the measured torque, the modeled torque showed a mismatch especially for high wrist flexion angles. One potential source of the model error is the inaccuracy of the endpoint tracking, as it is challenging to attach the markers directly to the end of the fPAM body close to the end-sealing knot. Another potential source of error is the inaccuracy regarding our method of mapping the 2D model to the human arm, as this mapping is not straightforward. We assumed that the marker positions projected onto the horizontal plane correspond to the planar model, but it is challenging to ensure that the flexion/extension muscles are actually along this plane and that the axis of rotation of the wrist is perpendicular to it. The mapping should be further improved by measuring the rotational axis of the biological wrist joint (e.g., with the use of more markers) and redefining the mapping plane based on it.

The second method of calculating the modeled torque, which includes an endpoint stretching model, allowed us to use the motion capture data only to derive the initial values of the placement parameters. This method showed reduced modeling error (27.7\% average error compared to the 34.3\% error of the former method), especially for high wrist flexion angles (Fig.~\ref{fig:ExperimentalData}(a)).

Our future work focuses on using this method to provide torque estimation during the operation of the exosuit. First, we aim to develop a parameter identification method to derive both initial placement parameters and stretching coefficients through a calibration process using only onboard sensors.
Also, further work can be done to improve the torque model. Regarding the endpoint displacement, the force-displacement relationship should be examined in more detail to confirm the general applicability of the presented formula or add improvements to it. Also, the single wrist radius parameter is an oversimplification of the geometry of the wrist surface. One way to improve our approximation of the wrist surface is to use two different radii for the distance from the wrist surface to the center of rotation of the wrist and the center point for the radius of curvature. Another option is to use more complex, non-circular geometric description or consider data-based, user-specific models as, for example, in~\cite{neuman2023user}.

The results of the torque measurement (Fig.~\ref{fig:ExperimentalData}(a)) additionally show that, despite reaching a peak torque of 3.3~Nm, the torque of the exosuit prototype is smaller than the optimized torque (Fig.~\ref{fig:ParameterOptimizationResult}), which highlights the challenges of fabricating the fPAM with the desired parameters ($L_0$, $r_0$, $\epsilon_{max}$), interfacing the exosuit with the human upper limb. This latter includes the issue of the measurement setup covering mounting surfaces and the change of the position of the mounting points during operation due to stretching.

\subsection{Results of the range of motion measurement} \label{subsec:Discussion_RoM}

\textbf{The fPAM-actuated exosuit has a restricted range of motion.} In comparison with the biological range of motion, we expected to get a reduced exosuit-actuated range in flexion, as the measured torque profile showed that the flexor fPAM can not apply torque to move the wrist over approximately 50$^\circ$. The results (Table~\ref{tab:RoM}) confirmed the reduced range, which was 44.5$^\circ$ compared to the biological 80.9$^\circ$. In the case of wrist extension, the exosuit-actuated range was smaller than the biological (38.7$^\circ$ compared to 56.0$^\circ$), although proportionally it is closer to the biological range than the exosuit-actuated range for flexion. The range was smaller than the biological for ulnar/radial deviation as well (42.7$^\circ$ compared to 62.3$^\circ$). This result was unexpected because the reduced range due to the increased $w_2$ distance (compared to the same distance in flexion/extension) was compensated by moving the mounting points closer to the wrist (decreased $d_2$).

We also found a mismatch between the measured and the modeled joint limits in all four directions of motion, with the smallest error for the flexor fPAM (Table~\ref{tab:RoM}). This highlights the need to identify the parameters of the fPAM in use (the parameters used to model all the fPAMs were those of the flexor fPAM). Besides the fPAM parameters, more accurate endpoint tracking and model mapping (as discussed for the torque measurement) are also necessary to get accurate placement parameters for the RoM prediction.

\textbf{The limited range highlights the importance of restricting the endpoint stretching and increasing the maximum fPAM contraction.} For the current exosuit design, the refinement of the placement on the body and the reinforcement of the material of the glove and elbow band at the mounting points can increase the range of motion in all directions. In the case of wrist flexion, to reach the full biological range of motion, we need to find ways to increase the maximum contraction ratio of the fPAM. One solution to slightly improve the maximum contraction of the fPAM is to use it in a series pneumatic artificial muscle configuration~\cite{greer2017series}, where constrictions are placed at regular intervals along the actuator. Also, finding an alternative routing for the fPAMs in which we can use longer muscles could increase the absolute length change. A potential drawback of this solution, however, is that the exosuit could become more bulky.

\subsection{Results of the control and demonstrations}

\textbf{The introduced antagonistic feedback control effectively demonstrates the exosuit's capacity for achieving desired wrist configurations, but it should be further improved to adapt to quick changes.} In the case of the planar positioning task (Fig.~\ref{fig:Demonstration_positioning}(b)), the actual joint angle reached the increased or decreased desired angles with a significant delay and overshoot. When the joint angle was increased, the desired position was reached with 0.84~s rise time and  19$\%$ overshoot. When the angle was decreased, the goal position was reached quicker with 0.54~s rise time, but the overshoot increased to 44$\%$. By using pressure feedback, the unit change of pressure causes a different change of the torque depending on the actual wrist angle, so the rate of change in the joint angle is not well regulated. This analysis indicates that the applied feedback control with a constant gain needs to be improved to better adapt to such quick changes. Based on the temporal and spatial resolution of the IMUs (100 Hz sampling frequency, $\pm$2.5$^\circ$ heading accuracy of the magnetometer) and the current loop rate of 0.014~s, we assume that the system architecture does not contribute significantly to the tracking error and it can remain unchanged when improving the control.

To improve the control in our future work, other control architectures for pneumatic artificial muscle-actuated and cable-driven upper limb exosuits provide guidelines. As presented in~\cite{andrikopoulos2015motion}, an advanced nonlinear PID controller was applied for similar positioning tasks for a pneumatic artificial muscle-actuated wrist exosuit that provided significantly smaller tracking error. In this exosuit design, the mounting points were fixed to a rigid structure, but applying this feedback method could potentially improve the trajectory tracking performance of our device. For the cable-driven wrist exosuit in~\cite{li2020bioinspired}, the desired wrist flexion angle was controlled based on an inverse kinematic model. Although the tracking accuracy was not evaluated in a way that can be compared with our results, this approach highlights the benefit of adding a model-based feed forward term, which can be implemented if the kinematic model parameters are identified for our model. This resolves the shortcoming of the pure feedback control that it does not account for different changes in force as for a given pressure change at different contraction levels.

To execute dynamic tasks, control strategies such as gravity compensation~\cite{little2019imu, ghosh2022design, mukherjee2023adaptive} and admittance control~\cite{lotti2020adaptive, chiaradia2021assistive, lotti2023soft} have been used. For example, in~\cite{chiaradia2021assistive}, admittance control was applied to assist flexion of the wrist with a cable routed similarly as our flexor fPAM. Such a control could be transferred to our exosuit if a sensor is added to measure the fPAM force. Other promising results for accurately controlling pneumatic artificial muscle-actuated robotic systems, e.g., applying an energy-based nonlinear control method~\cite{liang2021energy} to realize accurate positioning, could also be applied to our device.

\subsection{Proposed movement assistance applications}

\begin{figure}[tb]
    \centering
    \includegraphics[width=\columnwidth]{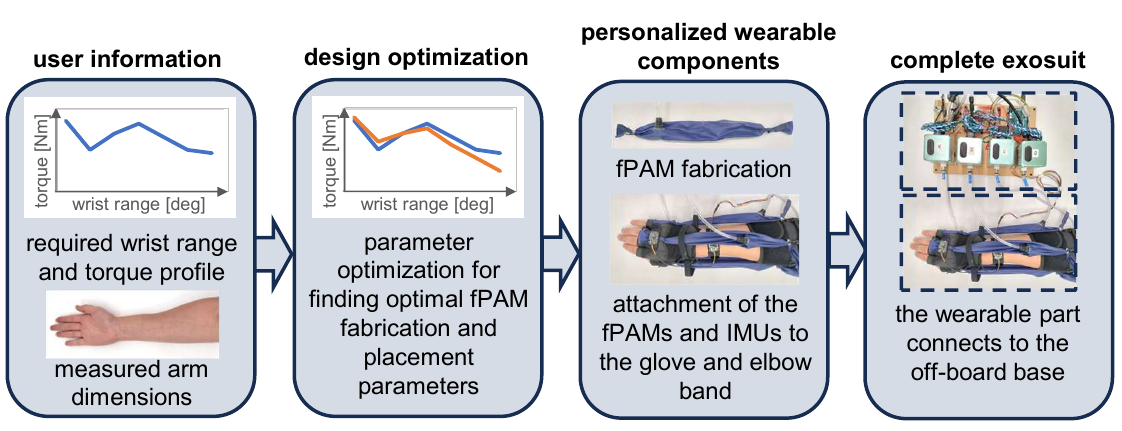}
    \vspace{-0.6 cm}
    \caption{Proposed pipeline to personalize the exosuit for a given user. User information is used to define the reference torque profile and the limits for the muscle placement parameters. The design optimization provides the optimal parameters according to the user's requirements, which are then used for the fabrication of the wearable components. Finally, the personalized wearable parts connect to the off-board base to complete the exosuit.}
    \vspace{-0.3 cm}
    \label{fig:Personalization_pipeline}
\end{figure}

\textbf{The exosuit could be personalized and used at home.} The wearable part of the exosuit is easy to fabricate and personalize for each user with the proposed pipeline shown in Fig.~\ref{fig:Personalization_pipeline}. The personalization considers the dimensions of the user's arm and the specific assistive torque that they require from the exosuit. Then, the optimization results provide the necessary fPAM lengths and attachment sites on the glove and forearm for fabrication. Through further research, we aim to examine how this process applies to users with different upper limb impairments and adjust the proposed process accordingly. The cost of the wearable part of the exosuit is low (approximately $\$$134). The off-board base components have a significantly higher cost (approximately $\$$3572 including a compressed air source), however, the same base can be used sequentially by multiple users. Also, the cost of all the components can be reduced, especially the pressure regulation components, which are the most expensive part of the base (approximately $\$$3229).

\textbf{Our primary proposed application is for conducting at-home rehabilitation exercises.} The base is compact enough to bring the device home, along with a portable air compressor or pump, and it can remain stationary when performing the exercises. Due to the limited fPAM stretching and contraction, the exosuit can be designed to have hard-stop limits, which makes it safe to operate without the supervision of a professional. Also, with feedback control, the exosuit is able to perform exercises where the wrist is slowly moved along a predefined trajectory. For stretching exercises, the fPAM is particularly promising, as it can produce high force, and the antagonistic configuration of the actuators is beneficial, as the stiffness can be gradually changed by regulating the co-contraction. Here, we only demonstrated the use of the exosuit for passively moving the wrist, but it could likely provide assistance and resistance as well. The observed limited torque and movement range in wrist flexion, however, restricts the usefulness of the exosuit for stretching across the full range of motion, therefore, further research should focus on overcoming this limitation as discussed in Section~\ref{subsec:Discussion_RoM}.

\textbf{The improvement of the exosuit's portability and control is required to provide movement assistance for daily activities.} Currently, the base of the exosuit is not portable, so the user must be tethered while wearing the exosuit. Also, the implementation of more advanced control and sensing is necessary for this application, including intent detection such as by recording myoelectric signals~\cite{fu2022myoelectric}. The control should be improved to better utilize the quick dynamic response of the fPAM~\cite{NaclerioRAL2020} and produce a quicker exosuit response. The physical capabilities of the device, however, seem satisfactory in providing assistance, because people do not use the full range of motion for most activities~\cite{palmer1985functional}, and the required assistive torque for most activities is smaller than the peak biological torque that was used as a reference in this work.

\vspace{-0.1 cm}
\section{Conclusion} \label{sec:conclusion}

We presented a soft wrist exosuit which uses fabric pneumatic artificial muscles (fPAMs), soft actuators that previously have not been included in a wearable device, despite their promising features. The exosuit has four fPAMs in a symmetric arrangement to move the wrist in flexion/extension and ulnar/radial deviation.

We introduced a two-dimensional model of the fPAM placement to calculate the torque applied by the exosuit to the wrist, and we demonstrated the use of the model in design optimization to choose the muscle placement parameters to reach the reference peak biological torque. The results of an example optimization for a single user show that the exosuit torques can reach the biological peak torques, except for at high wrist flexion angles.

To validate the model, we measured the torque that a flexor fPAM applies to the wrist. We modeled the torque both by using fixed position parameters and by including a model of fabric stretching at the mounting points. Compared to the former method, the latter method increased the accuracy of the model, which highlights the importance of including stretching in the torque model. We also measured the biological and exosuit-assisted range of motion along the two degrees of freedom of the wrist, which confirmed the limited range primarily in wrist flexion. These results call attention to the need to restrict the endpoint stretching and increase the maximum fPAM contraction.

Still, with the current prototype, we demonstrated the capability of the exosuit to move the wrist, first, to a desired position in flexion/extension, and then to follow a desired trajectory in two degrees of freedom, using an antagonistic feedback control algorithm.

The results of the presented work showed that, compared to other pneumatic wrist exosuits, the fPAM actuator can be used to provide a strong, lightweight design that is currently limited in the range of motion. Potential applications include movement assistance (with improved control and portability) and rehabilitation (with improved range of motion). Our future work will explore how impaired users can benefit from using the device, and how we can improve the exosuit design and control to enhance its performance.

\section{Acknowledgements} \label{sec:acknowledements}
We thank Mark Plecnik for providing access to the laser cutting and tensile testing machines used for experiments. Also, we thank Nicholas Naclerio for useful discussions about fabric pneumatic artificial muscles and György Cserey, and Miklós Koller for useful discussions about the statistical analysis of the experimental results.
\vspace{-0.4 cm}

\bibliographystyle{IEEEtran}
\bibliography{library}

\end{document}